\RequirePackage{fix-cm}
\documentclass[twocolumn]{svjour3}          
\smartqed  
\usepackage{graphicx}
\usepackage{mathptmx}
\usepackage{upgreek}
\usepackage{url}
\usepackage{algorithm}
\usepackage{algorithmic}
\usepackage{fancyvrb}
\usepackage{rotating}
\usepackage{subfigure}

\journalname{Soft Computing Journal (Applications and Methodologies)}
\begin{document}

\title{Distributed Optimization in Wireless Sensor Networks:\\an Island-Model Framework
}


\author{Giovanni Iacca}

\institute{G. Iacca \at
	   INCAS\textsuperscript{3}, Innovation Centre for Advanced Sensors and Sensor Systems \\
	   P.O. Box 797, 9400 AT Assen, The Netherlands \\
           Tel~+31 (0) 592 860 000 \\
           Fax~+31 (0) 592 860 001 \\
           \email{giovanniiacca@incas3.eu}
}

\date{Received: date / Accepted: date}

\maketitle

\begin{abstract}
Wireless Sensor Networks (WSNs) is an emerging technology in several application domains, ranging from urban surveillance to environmental and structural monitoring. Computational Intelligence (CI) techniques are particularly suitable for enhancing these systems. However, when embedding CI into wireless sensors, severe hardware limitations must be taken into account. In this paper we investigate the possibility to perform an online, distributed optimization process within a WSN. Such a system might be used, for example, to implement advanced network features like distributed modelling, self-optimizing protocols, and anomaly detection, to name a few. The proposed approach, called DOWSN (Distributed Optimization for WSN) is an island-model infrastructure in which each node executes a simple, computationally cheap (both in terms of CPU and memory) optimization algorithm, and shares promising solutions with its neighbors. We perform extensive tests of different DOWSN configurations on a benchmark made up of $15$ 
continuous optimization problems; we analyze the influence of the network parameters (number of nodes, inter-node communication period and probability of accepting incoming solutions) on the optimization performance. Finally, we profile energy and memory consumption of DOWSN to show the efficient usage of the limited hardware resources available on the sensor nodes.
\keywords{Optimization \and Wireless Sensor Networks \and Computational Intelligence \and Distributed Computing \and Memetic Computing}
\end{abstract}

\section{Introduction}\label{sec:intro}
Wireless Sensor Networks (WSNs) is an emerging technology with potential groundbreaking applications in several fields of engineering, medicine, weather forecast, environmental monitoring, surveillance, disaster management, see for example \cite{bib:surveyWSN02}, \cite{bib:applWSN} and \cite{bib:surveyWSN08}. 
In essence, a WSN is a network of embedded sensing devices (also called \emph{motes}) endowed with communication capabilities, i.e. systems which are able 
to measure one or more physical quantities (e.g. temperature, humidity, light, etc.) and exchange information, through a protocol stack, with other entities in the network. Albeit extremely flexible and relatively cheap, wireless sensors pose strict hardware constraints due to their embedded, distributed nature. As a consequence, motes are usually small in size, and limited in terms of CPU power, memory, and energy. A smart usage of these resources is thus necessary to overcome these limitations and extend the lifetime - and the efficiency - of these systems.

According to Harrop and Das \cite{bib:IDTechEx_wsn}, the worldwide market of WSN ``will grow rapidly from \$0.45 billion in 2012 to \$2 billion in 2022''.
The potentialities of WSNs and the rapid growth of their global market make them extremely interesting both from an application and scientific point of view.
One of the first research areas being attracted to this world has been Computational Intelligence (CI) \cite{bib:ciSurveyWSN}. In the last decade, several applications of CI techniques (e.g. Evolutionary Algorithms, Genetic Programming, Swarm Intelligence, Neural Networks, Fuzzy Logic, Reinforcement Learning) have been proposed in the context of WSN, see for example \cite{bib:psoSurveyWSN} and \cite{bib:eaSurveyWSN} for a survey of applications of Particle Swarm Optimization and Evolutionary Algorithms.

However, most of the existing works in the field focus on problems that can be solved offline on a centralized system, for instance optimal deployment, localization and clustering. To the best of our knowledge, very limited work has been done on distributed online CI approaches: for example, \cite{bib:gpWSNb}, \cite{bib:idgpWSN} and \cite{bib:gpWSNa} focus on the application of distributed Genetic Programming on WSNs; Rabbat and Nowak \cite{bib:RabbatNowak2004,bib:RabbatNowak2005} present instead a WSN-based distributed optimization framework for solving parameter estimation problems: given some assumptions on the fitness function, a parameter estimate is circulated through the network, and each node incrementally adjusts the estimate based on its local data. 

In line with these studies, in our previous work \cite{bib:DOWSN-UKCI} we introduced DOWSN, Distributed Optimization in Wireless Sensor Networks (pronounced ``dawson''), a decentralized, island-model framework designed to perform online optimization processes in a WSN. It should be noticed that, compared to the works presented in \cite{bib:RabbatNowak2004} and \cite{bib:RabbatNowak2005}, DOWSN does not require any specific assumption on the fitness function. In DOWSN, each sensor node executes a computationally cheap optimization algorithm and wirelessly exchanges, with a probability called \emph{imitation rate}, promising solutions with its neighboring nodes. Thanks to its flexible conceptual structure, this platform can be used, for example, to implement advanced WSN features such as distributed modelling, optimal scheduling of sensor readings, protocol optimization, etc. 

In this paper we extend the study of the DOWSN structure conducting an extensive campaign of numerical experiments and focusing in particular on the influence of the network parameters, namely number of nodes, inter-node communication period and imitation rate, on the global optimization performance. In addition, we perform a thorough analysis of energy and memory consumption, to show the efficient usage of the limited resources available on motes.
  
The rest of the paper is structured as follows. In the next section, we review briefly some previous studies on the application of CI in WSNs. The DOWSN architecture is described in details in section \ref{sec:dowsn}. Experimental results, obtained with different DOWSN configurations on the benchmark functions listed in appendix \ref{sec:test_desc}, are presented in section \ref{sec:numRes}, together with a detailed analysis of the influence of the network parameters on the optimization performance. In section \ref{sec:hardware} we report the profiling of energy and memory consumption of DOWSN. Finally, section \ref{sec:conclusion} concludes this work and suggests possible future developments.
\section{Related work}\label{sec:related_work}
Broadly speaking, the extant Computational Intelligence literature focuses on two classes of WSN applications, namely: (i) problems which can be solved offline, i.e before network deployment, by \emph{centralized} algorithms (e.g. optimal node placement, layout optimization); (ii) problems whose solution requires an online \emph{distributed} algorithm, involving node-local computations (e.g. energy-aware routing, localization, scheduling of measurements). In the following, we briefly review some selected works on these two classes of problems. It can be noticed how, mostly because of the aforementioned hardware constraints on the motes, while many studies have been done already on centralized offline applications, only few works actually focus on distributed online algorithms. Therefore, this area of research is still open.
\subsection{Optimal deployment}
Often it is needed that a WSN is deployed according to some optimality criteria depending on the position of motes (e.g. metrics measuring motes distribution, spatial coverage, network energy consumption, etc.). In static WSNs (i.e. networks where the position of motes is fixed), this problem can be efficiently solved offline prior to the actual deployment, simulating the network with one of the several simulation tools available \cite{bib:simSurveyWSNa}. Multi-objective optimization algorithms, such as Multi Objective Evolutionary Algorithms (MOEAs), can be used together with these tools to find optimal network configurations. Some studies on MOEAs applied to WSN deployment are reported in \cite{bib:ciSurveyWSN}, while the proper MOEA parameter setting for this problem is investigated in \cite{bib:moeaWSN}. An interesting algorithmic case is DPSOSA, a combined offline/online distributed hybrid algorithm presented in \cite{bib:dPSOsaWSN}: in DPSOSA, Particle Swarm Optimization (PSO) performs offline the 
global search, while Simulated Annealing (SA) is executed on motes to apply local refinements and online adjustments.
\subsection{Node localization}
Another issue which arises in WSNs (especially mobile networks) is finding the exact position of nodes. Similarly to optimal deployment, localization can be formulated as an optimization problem where the position error (difference between actual and estimated position) has to be minimized. 
In \cite{bib:ugaLocalizationWSN}, a micro-Genetic Algorithm is used to improve the accuracy of Ad-hoc Positioning System (APS), a WSNs-specific localization algorithm. An embedded implementation of a hybrid method combining the Gauss-Newton Algorithm (GNA) and the custom PSO \cite{bib:psoLocalizationWSN} is described in \cite{bib:rtLocalizationWSN}. Further applications of PSO and EAs are presented in \cite{bib:psoLocalizationWSN}, \cite{bib:eaLocalizationWSN}, while Kulkarni et al. \cite{bib:bioLocalizationWSN} propose a complete survey of bio-inspired techniques for localization in WSNs. An alternative approach is proposed in \cite{bib:Nguyen05}, where the localization problem is formulated in terms of pattern recognition and solved by means of a kernel-based learning algorithm.
\subsection{Clustering}
Clustering is a mechanism which creates (either static or dynamic) clusters of nodes, where a node is elected cluster-head and collects data from its neighbors. Optimizing the packet traffic, clustering is an effective means to balance energy consumption, improve network life-cycle and ensure a reliable communication.
Several clustering algorithms have been proposed in the context of WSNs. For example, in \cite{bib:clusterGAWSN} cluster-head selection is performed modeling the WSN as a neural network (being each node a neuron) whose structure and weights are adjusted by means of Genetic Algorithm (GA). A similar study is presented in \cite{bib:mopsoClusterWSN}, where Multi Objective PSO (MOPSO) is applied to dynamic clustering.
\subsection{Routing}
Yet another feature typical of WSNs which can be formulated as an optimization problem is routing, that is finding the best point-to-point path that network packets should follow. Depending on some physical/logical features of the network such as topology and spacial distribution of motes, different optimality criteria can be used for path selection, e.g. shortest path, minimum energy, maximum reliability, minimum packet loss, etc. In \cite{bib:evRoutWSN}, a cluster-based routing EA is presented where the fitness function incorporates a measure of cluster compactness and separation. In \cite{bib:routACOWSN}, an on-chip implementation of Ant Colony Optimization (ACO) for energy-efficient routing is described. Two examples of hybrid PSO-GA algorithms for energy-aware routing are instead presented in \cite{bib:gsoWSN} and \cite{bib:hEAWSN}. 
\subsection{Machine learning}
Machine learning represents an alternative means to improve network reliability and prolong network lifetime. An extensive survey of applications of machine learning in WSNs is reported in \cite{bib:mlWSN}. For example, in \cite{bib:decLearnWSN} a decentralized Reinforcement Learning (RL) is proposed, where each node is a self-learning agent whose purpose is finding the optimal schedule (i.e. active/sleep frames) that guarantees the highest energy efficiency and the minimum network latency. In \cite{bib:guestrin04c}, the optimal choice of sensor readings is performed combining a data model and live data acquisition, in order to guarantee the best balance between data accuracy and communication costs.
Another study \cite{bib:knowledgeWSN} proposes a fuzzy knowledge-based sensor network where each node infers information from its neighbors, thus providing a more accurate and reliable output. Distributed inference algorithms based on nonparametric models have also been proposed in \cite{bib:predd05a} and \cite{bib:predd05b}, while 
papers \cite{bib:guestrin04b}, \cite{bib:guestrin04a} and \cite{bib:guestrin05} describe probabilistic inference methods where nodes transmit, instead of raw data, constraints on the model parameters, thus drastically reducing the communication cost. 
\subsection{Automatic Programming}
One of the most recent trends in WSNs is the application of Genetic Programming (GP). Works like the aforementioned \cite{bib:gpWSNb}, \cite{bib:idgpWSN} and \cite{bib:gpWSNa} have started to investigate the possibility of generating, by means of genetic paradigms, the code running on motes. In particular, \cite{bib:gpWSNa} proposes a Distributed Genetic Programming Framework (DGPF) where various optimality criteria (such as energy consumption, memory usage and code size) are taken into account while generating the code for performing a given task. A similar framework, called Broadcast-Distributed Parallel Genetic Programming, is proposed in \cite{bib:gpWSNb}, where each node runs, independently, a lightweight GP process and asynchronously exchanges genetic material with its neighbors (island model). An island model distributed Genetic Programming engine, called \emph{In situ} Distributed Genetic Programming (IDGP), is also proposed in \cite{bib:idgpWSN}. One of the main conclusions of these studies is that, 
notwithstanding a potential risk of premature convergence, a framework combining GP and exchange of information endows the motes with robust self-adapting capabilities with regard to unpredicted changes of the local conditions and the surrounding environment.
\section{DOWSN: Distributed Optimization in Wireless Sensor Networks}\label{sec:dowsn}
As we have seen in the previous discussion, although the studies of applications of CI in WSNs are flourishing, very few distributed online algorithms have been investigated so far. However, there are specific problems that can be solved efficiently with this kind of algorithms. Here we focus in particular on problems which can be formulated as an online \emph{optimization} process, such as training of mote parameters, self-adaptation of protocols, dynamic data fusion, online clustering, etc. 

It is known that there exist several efficient population-based optimization algorithms, such as Evolutionary Algorithms, Particle Swarm Optimization, etc. One of the main problems of these algorithms is that they are usually computationally expensive, both in terms of memory and CPU power, because (a) they need to store and process a population of solutions; and (b) they generally perform CPU-hungry mathematical operations such as matrix transformations, gradient approximation, sorting, etc. 

Nevertheless, there are some specific algorithms, such as classic single-solution local search methods and more recent global search algorithms like nuSA \cite{bib:Xinchao2011} and 3SOME \cite{bib:Iacca20123SOME}, which demand less computational power and can be considered memory-saving. Despite low hardware requirements, some of these algorithms have proven extremely flexible and efficient, especially in domains like robotics and embedded control systems. Thus these methods seem suited also for WSNs, where the hardware limitations of motes impose constraints on memory and CPU usage. 

However, if on one hand the resources on a single mote are limited, on the other the computational power available on the whole network proliferates. This idea is at the basis of DOWSN \cite{bib:DOWSN-UKCI}, a unique framework for performing distributed, online optimization processes on board of a WSN. 

DOWSN exploits the distributed nature of WSNs based on an \emph{island model}. This model, originally investigated by Tanese \cite{bib:DGA87,bib:DGA89}, is a well-established distributed computing paradigm in the context of Genetic (and generally Evolutionary) Algorithms. According to the island model, each processing node executes a separate evolutionary algorithm (with a different initial population), while a periodical migration of individuals from one ``island'' to another is applied. For a comprehensive analysis of the island model in GAs, see \cite{bib:islandGA}. This model can be implemented naturally into a Wireless Sensor Network: in DOWSN, each mote executes independently a (memory-saving, rather than population based) optimization algorithm, and shares information, i.e. promising solutions, with it neighbors. Such a network can be seen, in a ``memetic'' metaphor, as an environment in which self-propagating units of information, or \emph{memes} (in this case, promising solutions) spread: in a 
cooperative approach, each agent shares its achievements with its neighbors, so that the best solutions can be forwarded to the whole network; on the other hand, in order to promote, at network-level, a beneficial diversity of ``ideas'' (which means diversity of solutions), a simple probabilistic mechanism which triggers the acceptance of incoming solutions is implemented at node-level (see below). The concept of population diversity, which is well-studied in the context of Evolutionary Algorithms, see e.g. \cite{bib:Zaharie2002b,bib:Zaharie2003a}, is thus applied here in the context of memes: preventing a detrimental homogeneity of solutions, but rather keeping a certain degree of solution diversity throughout the network, guarantees a richer pool from which new solutions are generated and, ultimately, a higher chance of finding the optimum.
\subsection{Hardware/Software architecture}
\begin{figure}[!ht]
\centering
\includegraphics[width=0.48\textwidth]{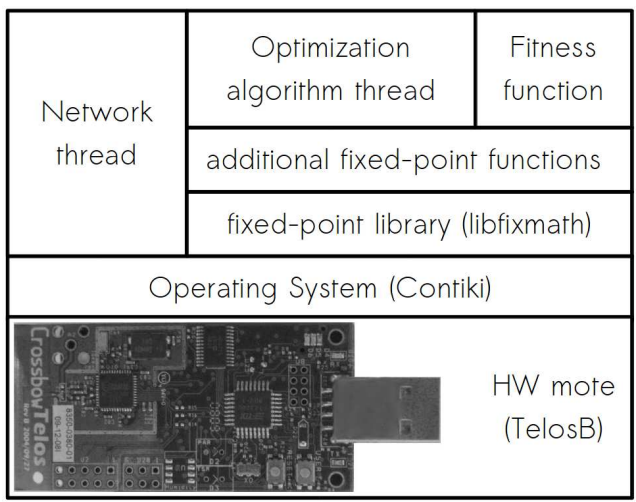}
\caption{Hardware/Software architecture of DOWSN}
\label{fig:mote_scheme}
\end{figure}

A schematic representation of the software architecture of DOWSN is shown in Fig. \ref{fig:mote_scheme} \cite{bib:DOWSN-UKCI}. Our reference hardware is the TelosB mote family \cite{web:openWSN}, an open-source platform endowed with a 8 MHz TI MSP430 micro-controller (16-bit RISC), 10 kB RAM, 48 kB program flash memory, 1 MB data flash memory, a CC2420 2.4 GHz IEEE 802.15.4 radio-frequency transceiver, and a sensor suite including light, temperature and humidity sensors. Different mote families can also be used (e.g. MicaZ/Mica2, Iris, Imote 2.0, etc. see \cite{web:wsnNodes}), provided an adequate API for programming the mote.

The software on a mote consists of a bottom-up structure organized as follows. The underlying level is the Operating System (OS): to cope with the limited hardware resources of motes, we use Contiki \cite{web:contiki}, an ad hoc OS characterized by lightweight memory structures and simple scheduling mechanisms. Keeping the same conceptual framework, it is also possible to port DOWSN to other WSN-specific Operating Systems \cite{bib:osSurveyWSN}, for instance TinyOS \cite{web:tinyOS}.

On top of Contiki, we use \texttt{libfixmath} \cite{web:libfixmath}, an open-source cross-platform C library which allows for fast fixed-point maths in Q16.16 notation\footnote{The Q16.16 notation indicates that 16 digits are used for the fractional part and 16 for the integer part of the number. The representable range is $[-32768.0,32767.999985]$, with a precision of $1/65536=0.000015$. \texttt{libfixmath} provides an overflow detection mechanism which allows developers to check the correctness of operations.}. An additional set of mathematical functions (such as \texttt{abs()}, \texttt{min()}, \texttt{max()}, \texttt{pow()}, \texttt{log()} and \texttt{nth\_rooth()}) is also implemented to ease the development of the optimization algorithms described in the following. Thanks to this library, it is possible to overcome some of the limitations due to the lack of a Floating Point Unit (FPU) which characterizes the TelosB mote family (and other families as well): without it, floating-point programming (which 
is essential in continuous optimization algorithms) would be less immediate.

The fitness function, implemented in fixed-point maths, represents obviously the goal of the optimization process: without loss of generality, we refer to the minimization problem of an objective function $f\left(\mathbf{x}\right)$, where the candidate solution $\mathbf{x}$ is an array of $n$ continuous design variables in a decision space $\mathbf{D}$. Depending on the specific problem at hand, the fitness function might be for instance a cost function, the error of a model dynamically trained on the mote, an optimality criteria related to the mote behaviour or to the network protocol, the localization accuracy, and so forth. In this study, we test DOWSN using $15$ benchmark functions commonly used to test global continuous optimization algorithms (see appendix \ref{sec:test_desc}). It is important to note that the use of fixed-point maths imposes a limited representable range and a rather coarse precision, which in turn reflects in additional bounds on the search space and the fitness function. 

The core component of DOWSN, executed on each sensor node, is finally represented by two treads: the first running the optimization algorithm (pseudo-code \ref{alg:structure}); the second handling the periodical communication (pseudo-code \ref{alg:networkThread}). This design choice nicely decouples the two long-running operations and guarantees a more efficient event handling. 

As described in our previous work \cite{bib:DOWSN-UKCI}, the optimization algorithm executed on a node is selected from a lightweight collection of memory-saving algorithms. In what follows, we call this collection Algorithm Database (A-DB). While the generic algorithmic structure of the algorithms present in the A-DB is the same, the search logics adopted by each algorithm is different. Four algorithms are included in the A-DB, namely:
\begin{itemize}
  \item Random Search (RS), a purely stochastic global search \cite{bib:Brooks1958} which explores the search space iteratively evaluating a randomly sampled solution and replacing the current solution only if the new solution outperforms it.
  \item Intelligent Single Particle Optimization (ISPO) \cite{bib:ISPO}, a ``degenerate'' PSO which employs a single n-dimensional particle $\mathbf{x}$ rather than a swarm. At the beginning of the optimization, the particle is randomly sampled in the search space $\mathbf{D}$. Then each step of the algorithm consists of the following. For each i-th variable of $\mathbf{x}$, a learning factor $L$ is initialized to zero. Then the i-th variable is perturbed $H$ times: for each t-th perturbation ($t=1,2,\dots~H)$, a velocity factor $v$, computed as: 
  \begin{equation}\label{velocity}
  v=A/t^P \cdot rand(-0.5,0.5) + B \cdot L^t
  \end{equation}
  is added to the previous value of the variable, as in a standard PSO (${x_i}^{t+1}={x_i}^t+v$). $A$, $P$, and $B$ are the acceleration, acceleration power factor, and learning coefficient. After the i-th variable is perturbed, the fitness of the perturbed particle is calculated and compared with the fitness prior to the perturbation. If an improvement has been achieved (or the perturbed solution has the same fitness of the original particle), the learning factor is updated as $L=v$, otherwise it is reduced by means of a shrinking factor $S_f$: $L=L/S_f$. If $L$ becomes smaller than a precision value $\varepsilon$, then it is reinitialized to zero.
  \item non-uniform Simulated Annealing (nuSA) \cite{bib:Xinchao2011}, a SA variant which dynamically adapts the radius of the neighborhood from which trial solutions are sampled. At the beginning of the optimization, the radius is as big as the whole search space, while in later stages it is focused on the most promising area. During each k-th iteration, $N_s$ trial solutions are sampled into a neighborhood of the current solution $\mathbf{x}$, according to the following non-uniform perturbation:
  \begin{equation}\label{mutation}
    {x_i}'= \left\lbrace  { 
    \begin{array}{ll}
      {x_i + \Delta(k,U_i-x_i)} & {\mathrm{if}~\eta=+1} \\ 
      {x_i - \Delta(k,x_i-L_i)} & {\mathrm{if}~\eta=-1} 
    \end{array} }\right. 
  \end{equation}
  where $L_i$ and $U_i$ are the lower and upper bounds of the i-th variable and $\eta$ is a discrete random variable with values in $\{-1,+1\}$. The function $\Delta(k,y)$ returns a value in the range $[0,y]$ which approaches to zero as $k$ increases:
  \begin{equation}\label{delta}
    \Delta(k,y)= y\times(1-\rho^{(1-\frac{k}{N})^b})
  \end{equation}
  where $\rho$ is a uniform random number in $\mathcal{U}(0,1)$, $N$ is the maximum generation number, and $b$ is a parameter affecting the dependency of the neighborhood size on the iteration number $k$.
  \item 3 Stage Optimal Memetic Exploration (3SOME) \cite{bib:Iacca20123SOME}, a recently proposed Memetic Computing approach characterized by a sequential structure composed of three memes, named long, middle and short distance exploration, arranged so to have an increasing exploitation pressure. Similar to a random search, the long distance exploration samples a new trial solution $\mathbf{x_t}$ within the entire decision space; however, in order to partially preserve the results found so far, the trial solution inherits a small portion of the current best solution (elite), by means of the exponential crossover typical of DE \cite{bib:NeriDEcDE2011}. This mechanism is repeated until a fitness improvement is found. As soon as the long distance exploration detects a new promising solution, and thus updates the elite, the middle distance exploration is activated. This second operator focuses the search in a hyper-cube centered on the current elite, sampling a given number of individuals within it. This 
mechanism is repeated as long as new improvements are found (and the hyper-cube is moved as well). Finally, when the middle distance exploration fails at finding new improvements, the short distance operator refines the search descending the basin of attraction. This refinement is done using a steepest descent deterministic local search algorithm inspired by \cite{bib:Hooke-Jeeves1961} and \cite{bib:Tseng2008}.
\end{itemize}
For each algorithm (except RS, which is parameter-less) we use the parameter setting suggested in the original paper. Thus ISPO is configured with $A=1$, $P=10$, $B=2$, $S=4$, $\varepsilon=10^{-5}$, $H=30$; nuSA with $b=5$ and $N_s = 3$; 
3SOME with $\alpha_e=0.05$, $\delta=0.2$, $k=4$, $\rho=0.4$ and budget for short distance equal to $150$ iterations (see \cite{bib:Iacca20123SOME} for further details).

All the algorithms in the A-DB are implemented as inlined C macros, to guarantee a faster execution time and a smaller memory overhead (a function call requires instead a context switch and a memory stack). In addition to that, the algorithms perform only in-place replacements (thus needing less memory slots). Here we use the term ``memory slot" to refer to an n-dimensional array (a candidate solution) of fixed-point numbers. It can be easily seen that the RS, ISPO and nuSA employ only two slots (one for the current best solution and one for the perturbed trial solution), while 3SOME needs one more slots to store the initial elite which is used for replacements in the short distance operator. Moreover, to force the exploration within the search space, but also to prevent overflow issues, a toroidal handling of the bounds is used. This means that, given an interval $\left[a,b\right]$, if $x_i=b+\zeta$, i.e. the i-th design variable exceeds the upper bound by a quantity $\zeta$, its value is replaced with $a+\
zeta$. A similar mechanism is applied for the lower bound.

As shown in pseudo-codes \ref{alg:structure}-\ref{alg:networkThread}, a flag is used to indicate that the optimization process is completed (based on or more stop criteria: here we use maximum computation time and maximum number of fitness evaluations). Until one of the stop criteria is met, the optimization thread updates the node-local best, i.e. the best solution known so far at node-level, whenever an improvement is found. This node-local best is shared, and accesses in a synchronized way, with the network thread. The latter periodically listens to incoming packets (containing promising solutions found by other nodes) and sends the node-local best individual to the neighbors. The communication mechanisms used rely on the network primitives provided by the OS: in this work we use the \emph{best-effort local broadcast} provided by RIME, a low-power lightweight protocol stack available in Contiki.
\begin{algorithm}[!ht]
\caption{Optimization algorithm thread} \label{alg:structure}
    \begin{algorithmic}
    \STATE flag = false
    \STATE initialize iteration counter $t=0$
    \STATE algorithm initializations
    \WHILE {$t < budget$}
        \STATE pause process (synchronize with network thread)
        \STATE perform algorithm iteration
        \STATE update node-local best (if needed)
        \STATE update iteration counter $t=t+1$
    \ENDWHILE
    \STATE flag = true
    \end{algorithmic}
\end{algorithm}
\begin{algorithm}[!ht]
\caption{Network thread} \label{alg:networkThread}
    \begin{algorithmic}
    \STATE initialize broadcast communication
    \WHILE {!flag}
	\STATE wait for thread period
	\STATE broadcast sending/receiving
	\IF{$packet~received$}
	  \IF{$f_{\mathrm{received~best}}<f_{\mathrm{local~best}}$}
	    \IF{$rand()<q$}
	      \STATE update node-local best
	    \ENDIF
	  \ENDIF
	\ENDIF
    \ENDWHILE
    \end{algorithmic}
\end{algorithm}
\begin{figure}[!ht]
\centering
\includegraphics[width=0.48\textwidth]{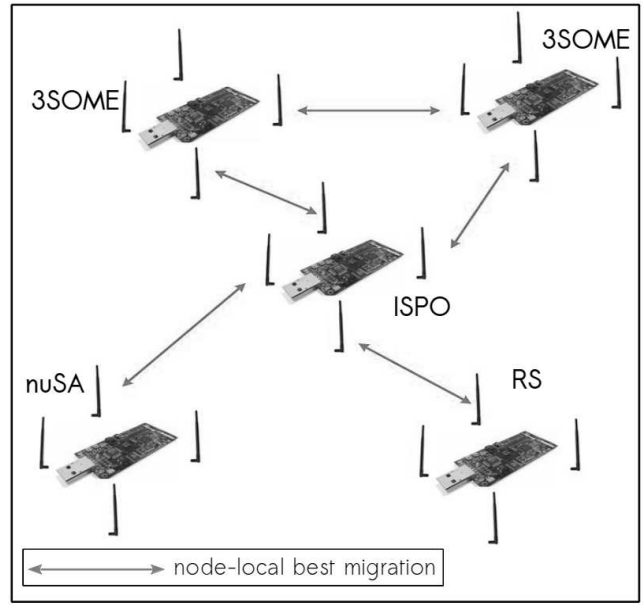}
\caption{Network-level scheme of DOWSN}
\label{fig:wsn_scheme_static}
\end{figure}
\begin{figure}[!ht]
\centering
\includegraphics[width=0.48\textwidth]{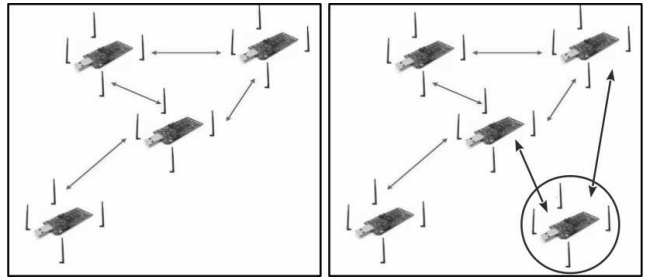}
\caption{Example scenario: a mote is added dynamically to the WSN}
\label{fig:wsn_scheme_dynamic}
\end{figure}

An important feature of DOWSN is the mechanism used to ``accept'' the incoming solutions, i.e. the way the node-local best is updated by the network thread. While the local improvements are always sent to the neighboring nodes, the incoming solutions replace the local best only with probability $q$ (of course in case of improvement). This parameter, named \emph{imitation rate}, can be interpreted similarly to the coefficient of imitation in the Bass diffusion model \cite{bib:BassDiffusion}, since it regulates the spreading of promising solutions.
If on one hand the migration of solutions quickly guides all the nodes in the network towards promising regions, on the other it is possible that it leads them to prematurely converge to the same local optimum, leaving some other promising regions of the search space unexplored. The effect of $q$ (see section \ref{sec:numRes}) thus balances this exploitation pressure: higher values of $q$ favor a faster diffusion of the best solutions (stronger exploitation), while lower values foster exploration.
\subsection{Heterogeneous vs Homogeneous DOWSN networks}
Two kinds of DOWSN networks can be envisioned, i.e. (a) heterogeneous, where each sensor node executes a different algorithm (selected from the A-DB described above, either randomly or according to some heuristics) and (b) homogeneous, where all the nodes execute the same algorithm. An example of heterogeneous network is shown in Fig. \ref{fig:wsn_scheme_static}, where two motes execute 3SOME, one nuSA, one ISPO and one RS. In section \ref{sec:numRes}, the two configurations will be analyzed in details, as well as the impact of network size and inter-node communication period (i.e. the period of the network thread) on the optimization results.
\subsection{Example scenario: dynamic configuration}
An interesting example of the dynamics of a DOWSN network is shown in Fig. \ref{fig:wsn_scheme_dynamic}, where a node is added to the network at runtime. Specifically, the left image illustrates a preexisting DOWSN configuration consisting of four nodes. As described above, each of the nodes executes an optimization algorithm and shares its incremental achievements with its neighbors. As soon as a new node is added to the network (right image), either because an existing node was previously switched off or because (in a mobile WSN) an external mote enters the broadcast range of the initial network, it automatically communicates with the other nodes and shares information with them. In case the new node has just been booted (meaning that an optimization process is just starting) it is quite likely that its current node-local best is not as good as the best solution known by the rest of the network. Therefore the new node will soon receive (and possibly accept) the best solution known by the network and it 
will continue its local optimization process to improve upon it, thus avoiding the exploration of less promising search regions. It is important to notice that this behaviour turns in a better use of computational power, which essentially means energy. A dual situation occurs if the new node knows a best solution which outperforms the best solution currently known by the other nodes in the network. This might happen, for example, in a mobile network where the new node is actually moving from a sub-network which already obtained a good solution to another which so far has been explored less promising areas of the search space. In this case the new node will share its information with the second sub-network and hopefully guide the algorithms running on it towards more promising parts of the decision space. 
\section{Numerical results}\label{sec:numRes}
In order to determine the individual contribution of the algorithms and the inter-node exchange of information to the overall performance, we define four different DOWSN configurations, namely:\begin{itemize}
 \item DOWSN-SA, a homogeneous DOWSN network composed of nodes executing 3SOME and exchanging information with their neighbors;
 \item DOWSN-SA stand-alone, a homogeneous DOWSN network as above, where the exchange of information is inhibited;
 \item DOWSN-MA, a heterogeneous DOWSN network composed of nodes executing an algorithm randomly selected from the A-DB and exchanging information with their neighbors;
 \item DOWSN-MA stand-alone, a heterogeneous DOWSN network as above, where the exchange of information is inhibited.
\end{itemize}
Here ``SA'' and ``MA'' stand for ``Single Algorithm'' and ``Multiple Algorithms'', respectively. We must remark that, for what regards the two DOWSN-SA configurations, 3SOME has been preferred to the other algorithms due to its generally better performance, as shown in \cite{bib:Iacca20123SOME}. Further experiments, not reported here for the sake of brevity, have also proven that DOWSN-SA configurations based on 3SOME tend to outperform analogous configurations based on one of the other algorithms present in the A-DB. 

In order to simulate complex optimization processes and assess the scalability of DOWSN, we consider the testbed described in appendix \ref{sec:test_desc} in three different problem dimensionalities, namely $5$, $15$ and $25$ variables\footnote{It should be noted that the packet structure used by the RIME protocol stack imposes an upper bound of $128$ bytes for the payload, which in turns limits the maximum amount of information that can be exchanged among nodes. Considering this limit, a maximum number of $128/4=32$ Q16.16 fixed-point values can be reliably transferred over RIME. Since each packet exchanged in DOWSN contains an n-dimensional array encoding an individual and its fitness (also in Q16.16 format), the upper limit for problem dimension in DOWSN is $31$. To overcome this limitation and handle solutions of higher dimensional optimization problems, an application-level protocol should be implemented on top of RIME.}. In the following, we assume a network global computational budget of $60$ seconds (
i.e. the optimization process is stopped after the timeout is exceeded) and $1000$ fitness evaluations per each node. Each DOWSN configuration is tested performing WSN simulations by means of COOJA \cite{bib:cooja2006}\footnote{COOJA is a cross-level simulator for Contiki which allows for simultaneous simulation at network, OS and machine code level. It includes several post-processing plugins, e.g. to estimate the power consumption on each node based on a simple energetic model.}. We assume also that each node boots at the beginning of the simulation (time $0$). For test purposes, whenever a fitness improvement is found we log on the standard output (saved as text file at the end of the simulation) the tuple \texttt{$\langle$timestamp, node id, fitness count, fitness value, solution$\rangle$}, as well as all the events captured by the network thread (sent \& received DOWSN packets)\footnote{In a real WSN deployment, these data might be collected on the data flash memory on the motes and then analyzed for 
post-processing. Another option would be a ``sink'' node connected to a PC: in this scenario, the sink would listen periodically to broadcast packets in order to provide the user, in real-time, the global output.}. These results are then post-processed by means of ad hoc Python scripts.
\subsection{Optimization performance}\label{sec:optimization}
In the first part of the experimentation, we focus on randomly generated network topologies composed of $5$ nodes, with an imitation rate of $0.9$ and a communication period of $0.25$ s. The effects of the network size, the imitation rate and the communication period are investigated in sections \ref{sec:size}, \ref{sec:imitation} and \ref{sec:period}, respectively. Each network is simulated $16$ times, each simulation being fed with a random seed generated externally from a Java random number generator. The number of simulations is chosen so that the average ``network'' fitness value at the end of the allotted budget is significant within a confidence interval $W=\sigma$, where $\sigma$ is the variance of the final ``network'' fitness value\footnote{Recalling that the standard error of the mean of a n-dimensional sample whose variance is $\sigma$ is $\sigma/\sqrt{n}$, and applying the central limit theorem to approximate the sample mean with a normal distribution, it follows that a sample size $n=16\sigma^2/
W^2$ 
guarantees a $95\%$ confidence interval of width $W$.}. By ``network'' fitness value, here we mean the global fitness value obtained by DOWSN, computed as average of the node-local best fitness values at the end of the budget. We use the average, rather than the minimum fitness, to have an indicator of the performance of the network as a whole. However, it should be noted that the global output in terms of problem solution should be defined differently, e.g. considering at any given moment the best individual among those ones found by all the motes composing the network.

Tables \ref{tab:benchmark05}-\ref{tab:benchmark25} show, for each test problem, the final ``network'' fitness value averaged over $16$ simulations and the corresponding standard deviation. The bold font indicates the best performance for each test function. To strengthen the statistical significance of the results, for each test problem we also report the outcome of the Wilcoxon rank-sum test \cite{bib:Wilcoxon1945}, applied with a confidence level of $0.95$, to compare the results of DOWSN-SA to those of the three other configurations. We indicate with ``='' an acceptance of the null-hypothesis (that the two DOWSN configurations under comparison are statistically equivalent from an optimization point of view), and with ``+'' (``-'') a superior (worse) performance of DOWSN-SA with respect to the configuration named as the label of the table column preceding the Wilcoxon column.

From Tables \ref{tab:benchmark05}-\ref{tab:benchmark25} it is clear that considering small-sized networks ($5$ nodes), a homogeneous configuration consisting of nodes using 3SOME and exchanging information with their neighbors (DOWSN-SA), outperforms on a regular basis homogeneous (3SOME-based) configurations where the inter-node communication is inhibited and heterogeneous DOWSN networks (with or without communication), i.e. networks employing different optimization algorithms. In particular, it can be seen that DOWSN-SA obtains the best results on $13$ test functions (out of $15$) in case of $5$ dimensions, and $12$ test functions in case of $15$ and $25$ dimensions. Additionally, DOWSN-SA finds the global optimum in six $5$-dimensional problems and one case ($f_{14}$) in $15$ dimensions. Remarkably, only in four cases (one in $5$ dimensions, one in $15$ and two in $25$), DOWSN-SA is statistically outperformed by another configuration (DOWSN-MA); in all the remaining cases, DOWSN-SA outperforms (or is 
statistically equivalent to) the other configurations.

Comparing only DOWSN-SA against its stand-alone variant, it is rather clear that the exchange of information (i.e. best individuals) is beneficial from an optimization point of view. The benefits of this exchange are particularly evident on higher dimensional problems ($15$ and $25$ variables), where the inter-node communication produces a fitness improvement in $14$ cases, while on low-dimensional problems ($5$ variables) an improvement is obtained only on five test problems. This might be explained with the relative simplicity of low-dimensional problems, which can be efficiently solved also by stand-alone optimization processes. Nevertheless, there are some heavily multimodal problems such as the Ackley ($f_3$) and Michalewicz ($f_6$) functions, for which even in $5$ dimensions the communication is able to produce a relevant fitness improvement. When the problem dimension increases (see Tables \ref{tab:benchmark15}-\ref{tab:benchmark25}), the positive effect of the inter-node communication is instead 
clear on all the test functions.

On the other hand, the fact that homogeneous networks tend to outperform heterogeneous ones deserves thoughtful considerations. One reason for this result might certainly be seen in the choice of the optimization algorithms used in the experiments. In other words, since a single (stand-alone) instance of an optimization process based on 3SOME generally is more successful \cite{bib:Iacca20123SOME} than an optimization based on ISPO, nuSA, or RS, it is likely that a network composed of all nodes running 3SOME is globally more efficient, from an optimization point of view, than a network including nodes executing different algorithms randomly chosen from the A-DB. This intuition is especially validated by the comparison DOWSN-SA vs DOWSN-MA stand-alone. The latter configuration, without the communication mechanism, is clearly penalized since some algorithms in the A-DB are not as efficient as 3SOME, thus producing a poorer network performance. This performance unbalancing is somehow mitigated when the inter-
node communication is activated (DOWSN-SA vs DOWSN-MA), as this mechanism allows for a rapid spreading in the network of the improvements obtained by the most efficient optimization algorithms. This means that in DOWSN-MA also the nodes running less efficient algorithms are able to exploit the improvements obtained by the other nodes in the network and explore the most promising areas of the search space. Despite this fact, however, DOWSN-MA shows a general performance slightly worse than DOWSN-SA, as it is clear from the comparison in Tables \ref{tab:benchmark05}-\ref{tab:benchmark25} where it can be seen that DOWSN-MA outperforms DOWSN-MA in $19$ cases (out of $45$), while as said is outperformed only in four cases. Yet it might possible that a heterogeneous network including memory-saving algorithms more efficient than 3SOME shows a higher performance than the DOWSN-MA configuration investigated in this work. To the best of our knowledge, however, the selected memory-saving algorithms (except the RS which 
was chosen only for testing purposes) are among the best memory-saving optimization methods available nowadays in literature, which makes 
difficult at the time of writing to envision better DOWSN-MA configurations.
\begin{table*}[!ht]
\small
\centering
\caption{Experimental results (average final value $\pm$ std. dev. and Wilcoxon test, reference DOWSN-SA) in $5$ dimensions. Homogeneous vs heterogeneous, distributed vs stand-alone networks with $5$ nodes.}
\label{tab:benchmark05}
\begin{tabular}{c|r@{$\,\pm\,$}l|r@{$\,\pm\,$}l|c|r@{$\,\pm\,$}l|c|r@{$\,\pm\,$}l|c}
\hline\noalign{\smallskip}
\# & \multicolumn{2}{c|}{DOWSN-SA} & \multicolumn{2}{c|}{DOWSN-SA stand-alone} & W & \multicolumn{2}{c|}{DOWSN-MA} & W & \multicolumn{2}{c|}{DOWSN-MA stand-alone} & W \\
\noalign{\smallskip}\hline\noalign{\smallskip}
$f_{ 1 }$ & $\mathbf{0.000e+00}$& $\mathbf{0.00e+00}$& $\mathbf{0.000e+00}$& $\mathbf{0.00e+00}$&=& $4.737e-06$& $1.48e-05$&=& $1.217e-01$& $6.21e-02$&+\\
$f_{ 2 }$ & $\mathbf{1.098e+00}$& $\mathbf{1.44e+00}$& $1.171e+00$& $6.90e-01$&=& $1.957e+00$& $1.36e+00$&=& $1.139e+01$& $4.49e+00$&+\\
$f_{ 3 }$ & $\mathbf{3.219e-04}$& $\mathbf{2.87e-05}$& $8.543e-02$& $1.41e-01$&+& $4.347e-02$& $1.37e-01$&+& $1.386e+00$& $4.57e-01$&+\\
$f_{ 4 }$ & $\mathbf{2.778e-04}$& $\mathbf{2.75e-05}$& $2.987e-04$& $3.40e-05$&=& $4.134e-04$& $9.23e-05$&+& $2.363e-02$& $8.51e-03$&+\\
$f_{ 5 }$ & $\mathbf{3.248e-03}$& $\mathbf{2.65e-04}$& $5.324e-02$& $8.62e-02$&+& $2.396e+00$& $2.41e+00$&+& $6.226e+00$& $7.39e-01$&+\\
$f_{ 6 }$ & $-4.009e+00$& $2.74e-01$& $-3.661e+00$& $2.38e-01$&+& $\mathbf{-4.029e+00}$& $\mathbf{2.62e-01}$&=& $-2.999e+00$& $1.77e-01$&+\\
$f_{ 7 }$ & $\mathbf{2.085e+03}$& $\mathbf{0.00e+00}$& $\mathbf{2.085e+03}$& $\mathbf{0.00e+00}$&=& $2.085e+03$& $3.49e-02$&=& $2.086e+03$& $2.60e-01$&+\\
$f_{ 8 }$ & $\mathbf{1.686e-05}$& $\mathbf{2.14e-05}$& $3.064e-05$& $1.61e-05$&+& $1.011e-03$& $2.25e-03$&=& $6.773e-02$& $3.50e-02$&+\\
$f_{ 9 }$ & $-2.000e+00$& $3.80e-04$& $-1.997e+00$& $8.77e-03$&+& $\mathbf{-2.000e+00}$& $\mathbf{0.00e+00}$&-& $-1.741e+00$& $7.54e-02$&+\\
$f_{ 10 }$ & $\mathbf{0.000e+00}$& $\mathbf{0.00e+00}$& $\mathbf{0.000e+00}$& $\mathbf{0.00e+00}$&=& $9.932e-03$& $1.03e-02$&+& $2.918e-01$& $9.71e-02$&+\\
$f_{ 11 }$ & $\mathbf{3.750e-07}$& $\mathbf{9.92e-07}$& $1.125e-06$& $1.80e-06$&=& $2.527e-04$& $2.76e-04$&+& $1.053e-01$& $5.23e-02$&+\\
$f_{ 12 }$ & $\mathbf{0.000e+00}$& $\mathbf{0.00e+00}$& $\mathbf{0.000e+00}$& $\mathbf{0.00e+00}$&=& $8.719e-05$& $1.25e-04$&+& $2.335e-01$& $1.00e-01$&+\\
$f_{ 13 }$ & $\mathbf{0.000e+00}$& $\mathbf{0.00e+00}$& $\mathbf{0.000e+00}$& $\mathbf{0.00e+00}$&=& $3.461e-04$& $5.52e-04$&=& $1.793e+00$& $8.50e-01$&+\\
$f_{ 14 }$ & $\mathbf{0.000e+00}$& $\mathbf{0.00e+00}$& $\mathbf{0.000e+00}$& $\mathbf{0.00e+00}$&=& $\mathbf{0.000e+00}$& $\mathbf{0.00e+00}$&=& $3.598e-02$& $2.96e-02$&+\\
$f_{ 15 }$ & $\mathbf{4.150e-06}$& $\mathbf{5.55e-06}$& $8.263e-06$& $6.85e-06$&=& $5.861e-03$& $5.38e-03$&+& $2.964e-01$& $1.23e-01$&+\\
\noalign{\smallskip}\hline
\end{tabular}
\end{table*}
\begin{table*}[!ht]
\small
\centering
\caption{Experimental results (average final value $\pm$ std. dev. and Wilcoxon test, reference DOWSN-SA) in $15$ dimensions. Homogeneous vs heterogeneous, distributed vs stand-alone networks with $5$ nodes.}
\label{tab:benchmark15}
\begin{tabular}{c|r@{$\,\pm\,$}l|r@{$\,\pm\,$}l|c|r@{$\,\pm\,$}l|c|r@{$\,\pm\,$}l|c}
\hline\noalign{\smallskip}
\# & \multicolumn{2}{c|}{DOWSN-SA} & \multicolumn{2}{c|}{DOWSN-SA stand-alone} & W & \multicolumn{2}{c|}{DOWSN-MA} & W & \multicolumn{2}{c|}{DOWSN-MA stand-alone} & W \\
\noalign{\smallskip}\hline\noalign{\smallskip}
$f_{ 1 }$ & $\mathbf{9.282e-04}$& $\mathbf{9.40e-04}$& $3.594e-03$& $1.08e-03$&+& $1.711e-03$& $1.55e-03$&=& $3.097e+00$& $7.15e-01$&+\\
$f_{ 2 }$ & $\mathbf{1.180e+01}$& $\mathbf{2.79e+00}$& $2.729e+01$& $8.31e+00$&+& $4.405e+01$& $2.86e+01$&+& $5.353e+02$& $1.09e+02$&+\\
$f_{ 3 }$ & $\mathbf{1.284e+00}$& $\mathbf{7.79e-01}$& $2.010e+00$& $7.76e-01$&+& $1.785e+00$& $1.02e+00$&=& $3.486e+00$& $4.61e-01$&+\\
$f_{ 4 }$ & $\mathbf{2.059e-03}$& $\mathbf{2.59e-04}$& $2.462e-03$& $2.62e-04$&+& $2.065e-03$& $3.28e-04$&=& $2.293e-01$& $5.25e-02$&+\\
$f_{ 5 }$ & $\mathbf{3.580e+00}$& $\mathbf{2.10e+00}$& $8.282e+00$& $4.53e+00$&+& $1.416e+01$& $8.40e+00$&+& $5.142e+01$& $7.46e+00$&+\\
$f_{ 6 }$ & $-8.032e+00$& $6.41e-01$& $-7.456e+00$& $5.20e-01$&+& $\mathbf{-8.416e+00}$& $\mathbf{8.69e-01}$&=& $-5.433e+00$& $6.55e-01$&+\\
$f_{ 7 }$ & $6.256e+03$& $2.75e-01$& $6.256e+03$& $3.62e-01$&+& $\mathbf{6.255e+03}$& $\mathbf{0.00e+00}$&-& $6.263e+03$& $1.50e+00$&+\\
$f_{ 8 }$ & $\mathbf{3.902e-01}$& $\mathbf{1.58e-01}$& $9.610e-01$& $1.76e-01$&+& $1.118e+00$& $5.88e-01$&+& $6.815e+00$& $1.65e+00$&+\\
$f_{ 9 }$ & $\mathbf{-1.918e+00}$& $\mathbf{6.35e-02}$& $-1.660e+00$& $1.19e-01$&+& $-1.801e+00$& $5.43e-01$&+& $-6.774e-01$& $2.59e-01$&+\\
$f_{ 10 }$ & $9.664e-02$& $3.81e-02$& $1.623e-01$& $4.46e-02$&+& $\mathbf{6.983e-02}$& $\mathbf{3.69e-02}$&=& $4.047e+00$& $7.01e-01$&+\\
$f_{ 11 }$ & $\mathbf{5.152e-03}$& $\mathbf{1.87e-03}$& $8.012e-03$& $1.75e-03$&+& $5.855e-03$& $2.68e-03$&=& $2.371e+00$& $4.83e-01$&+\\
$f_{ 12 }$ & $\mathbf{8.994e-03}$& $\mathbf{6.54e-03}$& $3.927e-02$& $2.24e-02$&+& $2.600e-02$& $4.50e-02$&=& $2.102e+01$& $5.00e+00$&+\\
$f_{ 13 }$ & $\mathbf{3.312e-02}$& $\mathbf{2.95e-02}$& $1.885e-01$& $1.05e-01$&+& $1.167e-01$& $9.88e-02$&+& $1.215e+02$& $2.12e+01$&+\\
$f_{ 14 }$ & $\mathbf{0.000e+00}$& $\mathbf{0.00e+00}$& $9.500e-07$& $3.68e-06$&=& $6.627e-05$& $1.61e-04$&=& $8.261e+02$& $2.17e+03$&+\\
$f_{ 15 }$ & $\mathbf{9.451e-01}$& $\mathbf{4.44e-01}$& $2.347e+00$& $4.51e-01$&+& $1.868e+00$& $8.13e-01$&+& $4.236e+02$& $1.58e+03$&+\\
\noalign{\smallskip}\hline
\end{tabular}
\end{table*}
\begin{table*}[!ht]
\small
\centering
\caption{Experimental results (average final value $\pm$ std. dev. and Wilcoxon test, reference DOWSN-SA) in $25$ dimensions. Homogeneous vs heterogeneous, distributed vs stand-alone networks with $5$ nodes.}
\label{tab:benchmark25}
\begin{tabular}{c|r@{$\,\pm\,$}l|r@{$\,\pm\,$}l|c|r@{$\,\pm\,$}l|c|r@{$\,\pm\,$}l|c}
\hline\noalign{\smallskip}
\# & \multicolumn{2}{c|}{DOWSN-SA} & \multicolumn{2}{c|}{DOWSN-SA stand-alone} & W & \multicolumn{2}{c|}{DOWSN-MA} & W & \multicolumn{2}{c|}{DOWSN-MA stand-alone} & W \\
\noalign{\smallskip}\hline\noalign{\smallskip}
$f_{ 1 }$ & $\mathbf{1.279e-01}$& $\mathbf{5.62e-02}$& $2.547e-01$& $8.60e-02$&+& $2.295e-01$& $3.88e-01$&=& $9.431e+00$& $2.06e+00$&+\\
$f_{ 2 }$ & $\mathbf{1.222e+02}$& $\mathbf{2.79e+01}$& $1.993e+02$& $2.88e+01$&+& $2.555e+02$& $8.42e+01$&+& $2.156e+03$& $4.94e+02$&+\\
$f_{ 3 }$ & $\mathbf{2.838e+00}$& $\mathbf{2.63e-01}$& $3.260e+00$& $2.26e-01$&+& $3.249e+00$& $4.76e-01$&+& $4.147e+00$& $2.89e-01$&+\\
$f_{ 4 }$ & $\mathbf{5.585e-02}$& $\mathbf{1.54e-02}$& $7.472e-02$& $1.33e-02$&+& $6.440e-02$& $4.76e-02$&=& $3.685e-01$& $6.32e-02$&+\\
$f_{ 5 }$ & $\mathbf{4.594e+01}$& $\mathbf{6.45e+00}$& $5.151e+01$& $3.57e+00$&+& $5.680e+01$& $1.20e+01$&+& $1.168e+02$& $1.45e+01$&+\\
$f_{ 6 }$ & $\mathbf{-7.783e+00}$& $\mathbf{5.86e-01}$& $-7.318e+00$& $7.55e-01$&=& $-7.321e+00$& $1.13e+00$&=& $-5.341e+00$& $5.43e-01$&+\\
$f_{ 7 }$ & $1.044e+04$& $1.33e+00$& $1.044e+04$& $9.90e-01$&+& $\mathbf{1.043e+04}$& $\mathbf{1.18e+00}$&-& $1.045e+04$& $2.45e+00$&+\\
$f_{ 8 }$ & $\mathbf{4.951e+00}$& $\mathbf{1.20e+00}$& $6.616e+00$& $5.60e-01$&+& $6.891e+00$& $2.63e+00$&+& $2.679e+01$& $7.52e+00$&+\\
$f_{ 9 }$ & $\mathbf{-1.460e+00}$& $\mathbf{2.56e-01}$& $-6.111e-01$& $2.90e-01$&+& $-8.426e-01$& $9.68e-01$&=& $-2.279e-02$& $2.54e-01$&+\\
$f_{ 10 }$ & $1.085e+00$& $2.69e-01$& $1.344e+00$& $1.42e-01$&+& $\mathbf{3.198e-01}$& $\mathbf{6.71e-02}$&-& $8.341e+00$& $1.50e+00$&+\\
$f_{ 11 }$ & $\mathbf{3.668e-01}$& $\mathbf{2.18e-01}$& $6.674e-01$& $2.40e-01$&+& $4.786e-01$& $4.53e-01$&=& $7.567e+00$& $1.21e+00$&+\\
$f_{ 12 }$ & $2.156e+00$& $8.77e-01$& $4.005e+00$& $9.51e-01$&+& $\mathbf{1.637e+00}$& $\mathbf{2.08e+00}$&=& $1.051e+02$& $2.15e+01$&+\\
$f_{ 13 }$ & $\mathbf{1.470e+01}$& $\mathbf{4.10e+00}$& $2.686e+01$& $8.49e+00$&+& $1.633e+01$& $1.92e+01$&=& $5.276e+02$& $9.64e+01$&+\\
$f_{ 14 }$ & $\mathbf{1.190e-01}$& $\mathbf{8.59e-02}$& $5.724e-01$& $4.11e-01$&+& $4.026e+01$& $9.10e+01$&+& $9.013e+03$& $4.42e+03$&+\\
$f_{ 15 }$ & $\mathbf{1.323e+01}$& $\mathbf{2.84e+00}$& $1.631e+01$& $2.07e+00$&+& $1.655e+01$& $2.99e+00$&+& $8.481e+02$& $2.16e+03$&+\\
\noalign{\smallskip}\hline
\end{tabular}
\end{table*}
\begin{figure*}[!ht]
\centering
\mbox
{
\subfigure[Michalewicz function, 5 dimensions]{\includegraphics[width=0.5\textwidth]{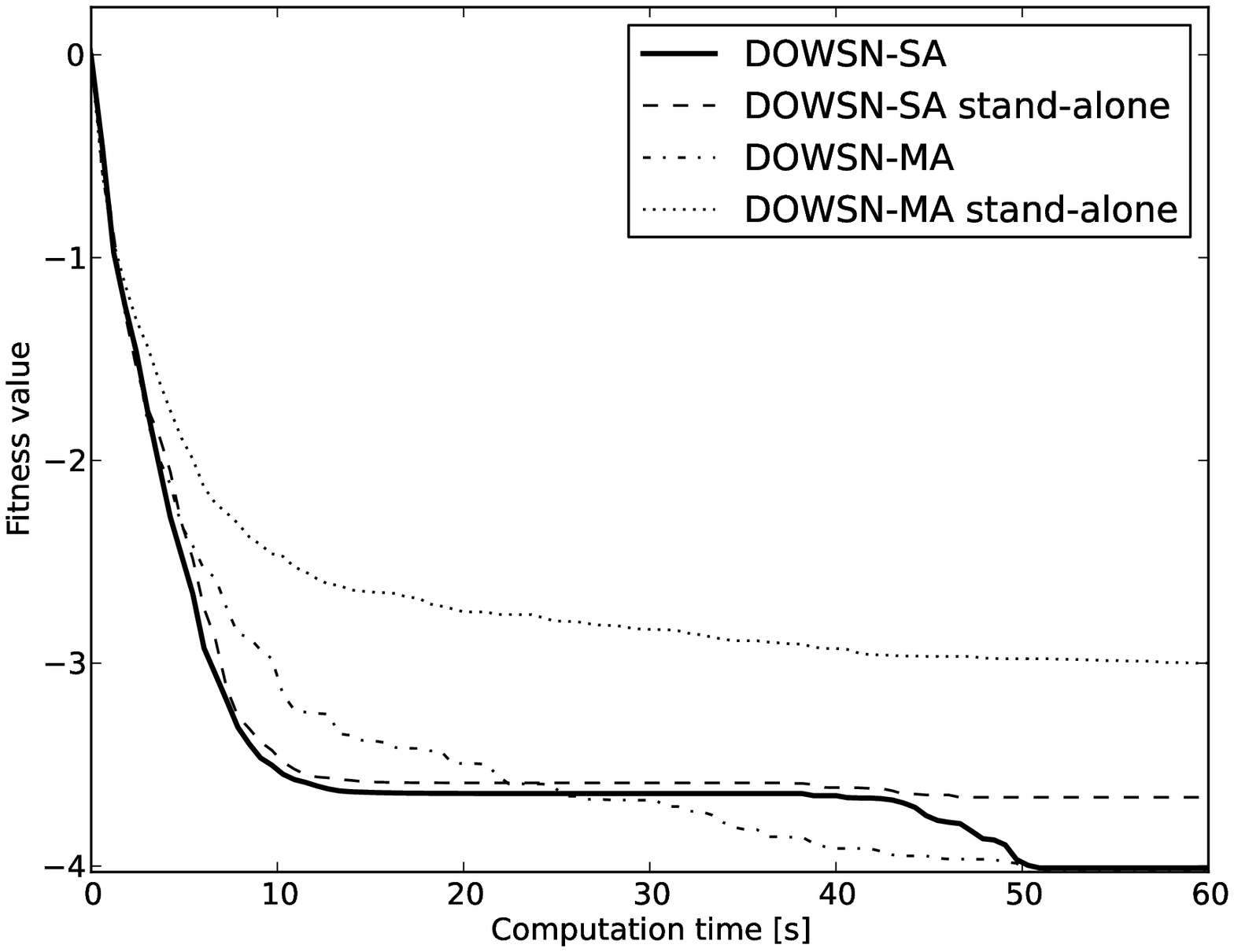}}
~
\subfigure[Schwefel function, 15 dimensions]{\includegraphics[width=0.5\textwidth]{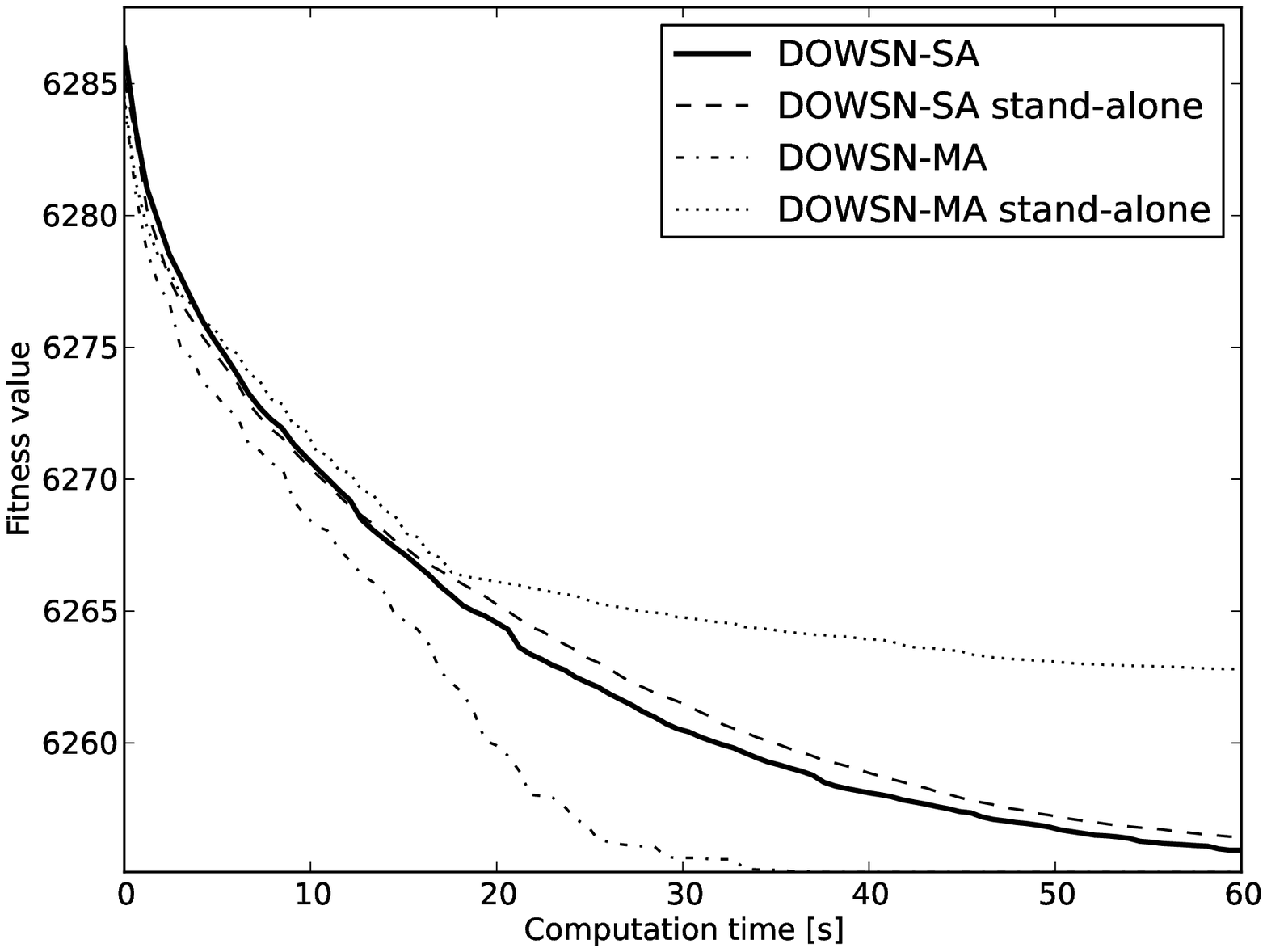}}
}
\mbox
{
\subfigure[Ackley function, 25 dimensions]{\includegraphics[width=0.5\textwidth]{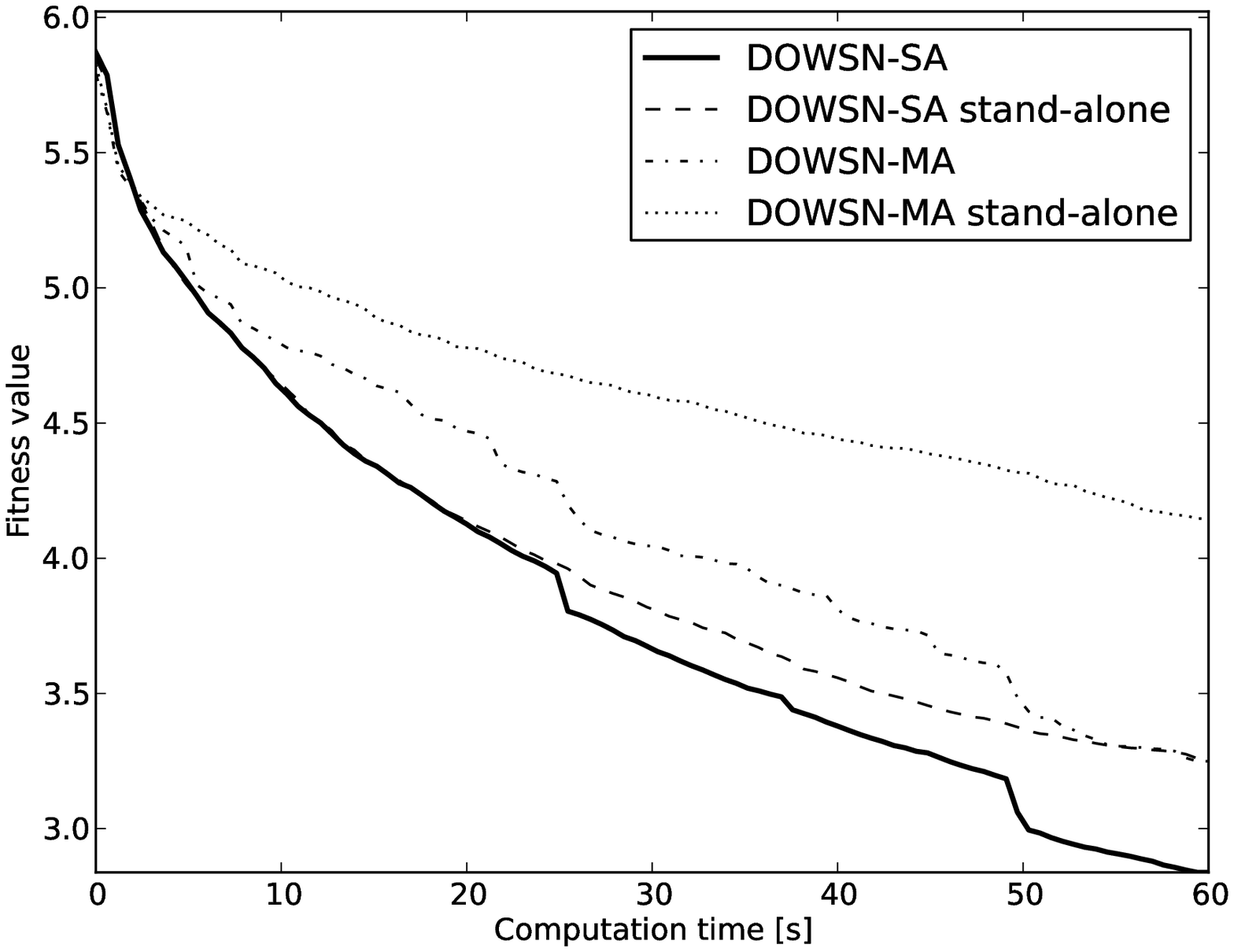}}
~
\subfigure[Schwefel problem 2.21, 25 dimensions]{\includegraphics[width=0.5\textwidth]{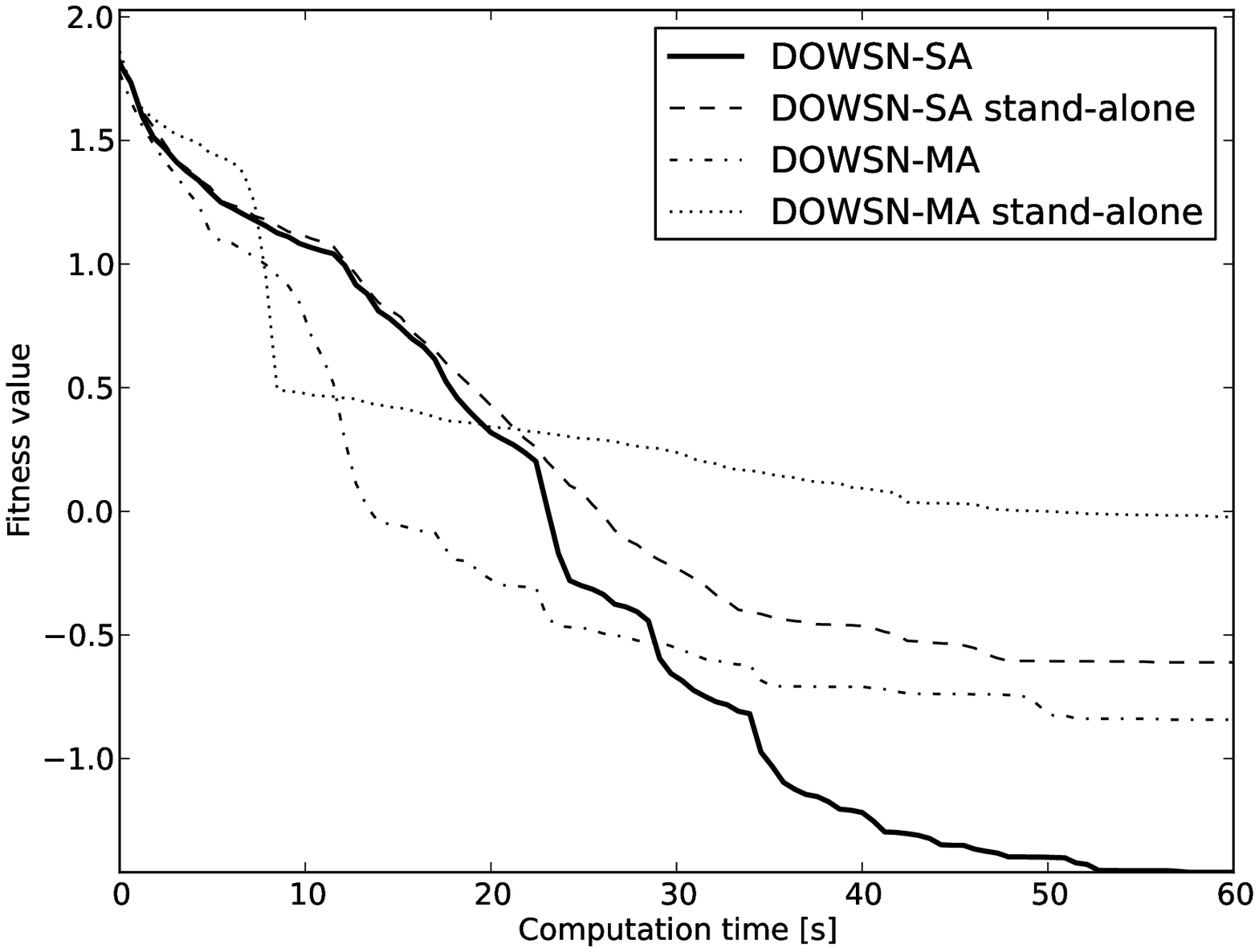}}
}
\caption{Average fitness trend obtained with four different DOWSN configurations (with 5 nodes)}
\label{fig:optimization}
\end{figure*}

A few examples of fitness trends (averaged over $16$ repetitions) obtained with the four aforementioned DOWSN configurations on four different test functions are shown in Fig. \ref{fig:optimization}. As reported previously, on $5$-dimensional problems one of the cases where the advantages of exchanging information among nodes is most clear is the Michalewicz function (see Fig. \ref{fig:optimization}.a), the reason being probably the high number of local optima ($n!$). On higher dimensional-problems, see Fig. \ref{fig:optimization}.c and Fig. \ref{fig:optimization}.d, the best performance of DOWSN-SA is even more evident, since it converges faster and to better final values compared to the other DOWSN configurations. Finally, Fig. \ref{fig:optimization}.b shows one of the four only cases (out of $135$ total comparisons) where DOWSN-SA is outperformed, in this case by the DOWSN-MA configuration.

\begin{figure*}[!ht]
\centering
\includegraphics[width=0.7\textwidth]{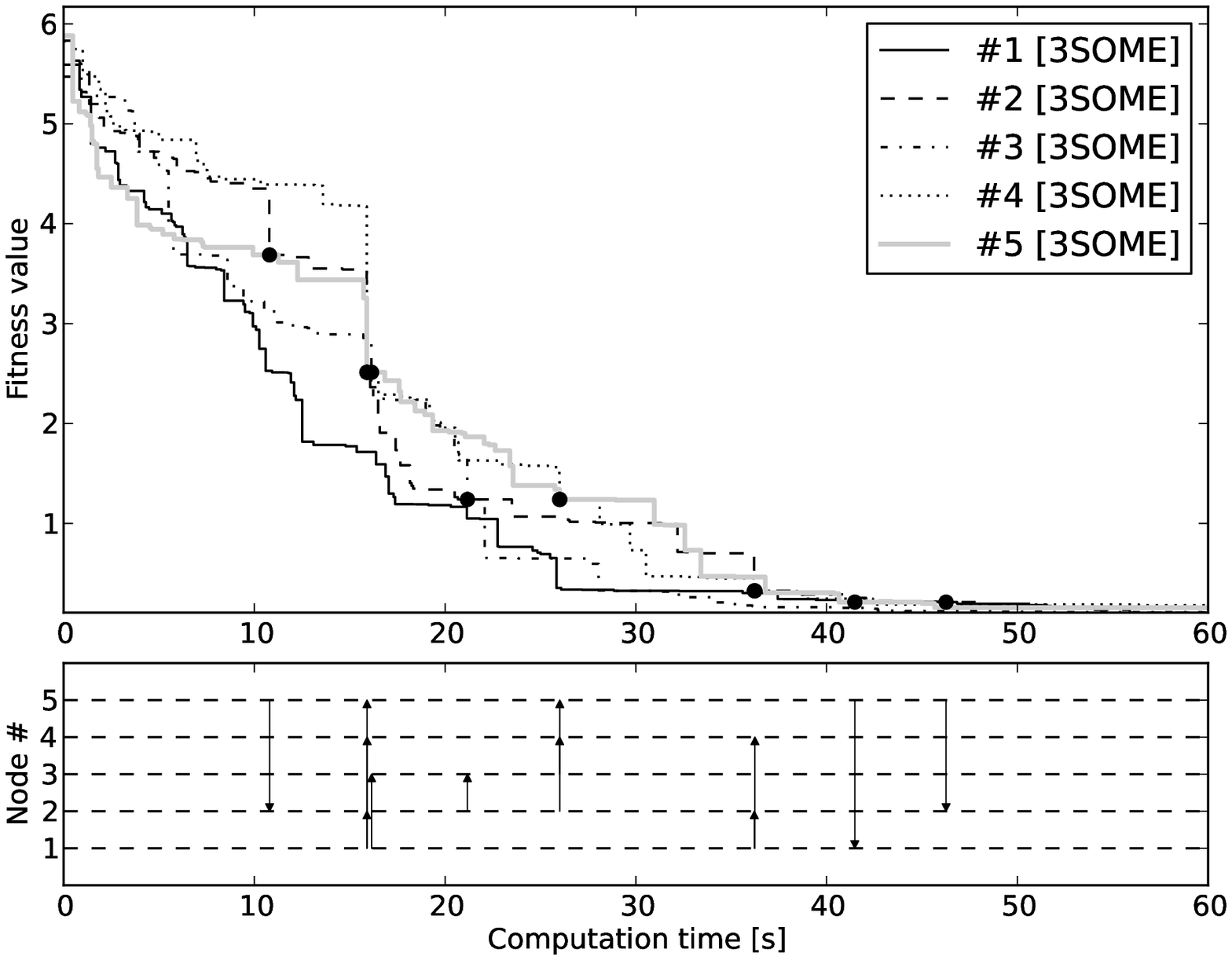}
\includegraphics[width=0.7\textwidth]{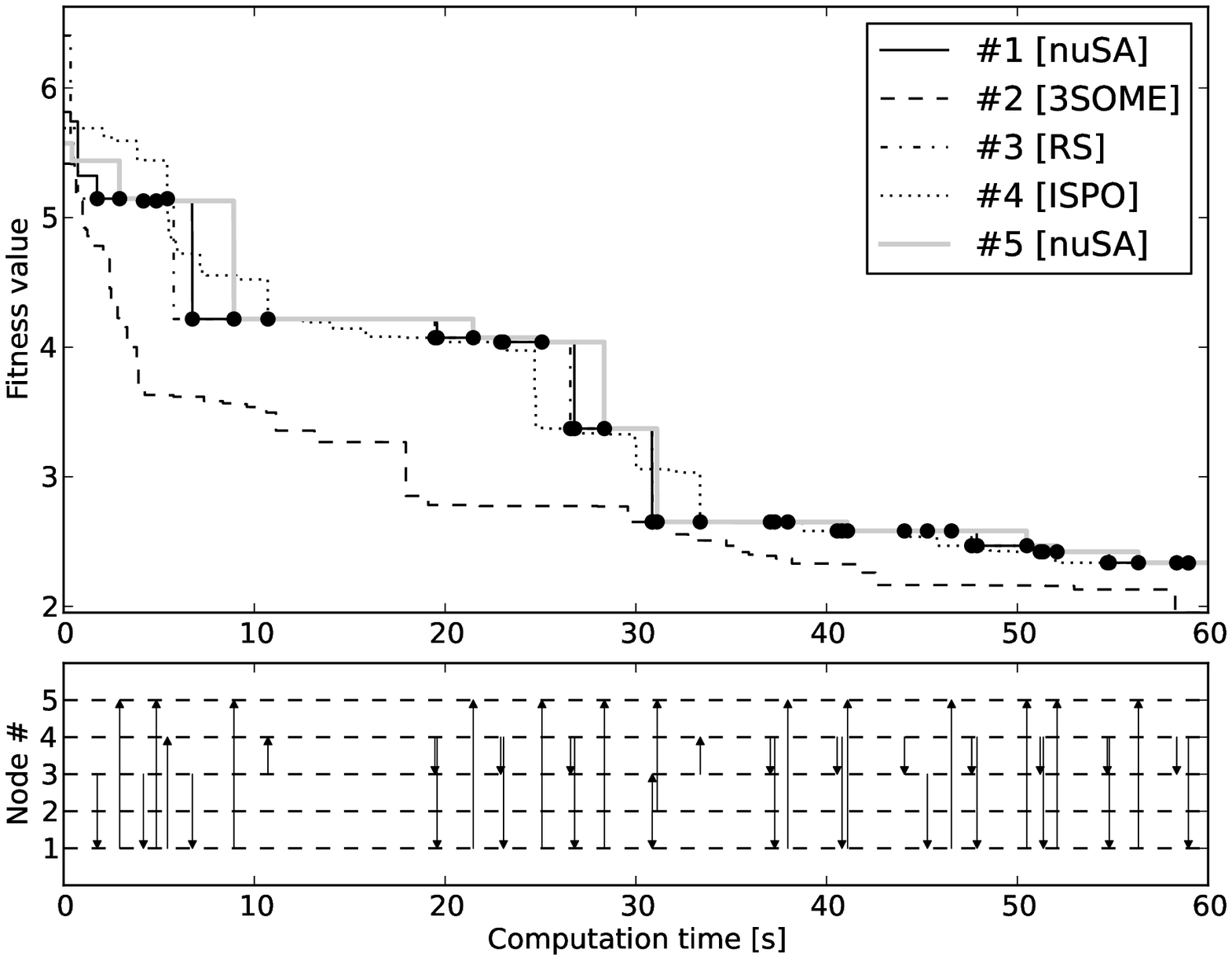}
\caption{Node-local fitness trends and network packet exchange obtained with DOWSN-SA (top) and DOWSN-MA (bottom), both with 5 nodes, on the Ackley function in 15 dimensions}
\label{fig:network_dynamics}
\end{figure*}

To conclude this discussion, we finally present some considerations about the network dynamics compared between DOWSN-SA and DOWSN-MA. Fig. \ref{fig:network_dynamics} shows the node-local fitness trends, together with the packet exchange, obtained during a single simulation of the two configurations (with the same parameter setting as before) optimizing the Ackley function in $15$ dimensions. The optimization algorithm executed on each node is shown in the legend next to the node id. In the example, due to the random selection of the algorithm in DOWSN-MA, it happens that two nodes run nuSA, one 3SOME, one ISPO and one RS. The black dots in the two topmost subplots represent a best individual update event caused by a node receiving an improvement from one of its neighbors. Each event is also represented in the two lowermost subplots as a small arrow from the node sending an individual to the one receiving it. It can be seen that these events obviously correspond to improvements in the fitness trend on the 
receiving node. Moreover, it is interesting to notice that although the total amount of network traffic (not shown in the figure) is almost the same for the two configurations, the heterogeneous network tends to produce more update events. This is mainly because the least efficient algorithms in the network receive frequent improvements from the most efficient ones, thus resulting in a higher number of update events. Conversely, in a homogeneous network the number of these events is lower, as the algorithmic dynamics on each node tends to be similar (despite the stochasticity of 3SOME). In other words the nodes show a similar search path and tend to converge to similar results. Still, the communication is useful also in a homogeneous configuration since on one hand it speeds up the overall convergence, on the other it may act as a ``disturbance'' (restart) mechanism, thus allowing for an improved exploration pressure. 

Recalling the ``memetic'' metaphor from section \ref{sec:dowsn}, the exchange of information among heterogeneous cultural entities, i.e. agents with different knowledge, has the beneficial effect of rapidly spreading worthwhile ideas to the whole network, naturally suppressing less promising ones. This turns eventually into a more frequent exchange of information. Instead, in a network where all the cultural entities have similar knowledge, while the chance for one agent to learn a novel idea (a fitness improvement) is lower, it may still happen either that the whole network reaches, by small incremental improvements, an optimum, or that, due to the random mechanisms behind the generation of new ideas, a breakthrough emerges.
\subsection{Influence of the network size}\label{sec:size}
\begin{table}[!ht]
\small
\centering
\caption{Experimental results (average final value $\pm$ std. dev. and Wilcoxon test, reference DOWSN-SA with 5 nodes) in 5 dimensions. Homogeneous DOWSN networks with 5 and 10 nodes.}
\label{tab:nr_nodes05}
\setlength{\tabcolsep}{.5em}
\begin{tabular}{c|r@{$\,\pm\,$}l|r@{$\,\pm\,$}l|c}
\hline\noalign{\smallskip}
\# & \multicolumn{2}{c|}{DOWSN-SA (5 nodes)} & \multicolumn{2}{c|}{DOWSN-SA (10 nodes)} & W \\
\noalign{\smallskip}\hline\noalign{\smallskip}
$f_{ 1 }$ & $\mathbf{0.000e+00}$& $\mathbf{0.00e+00}$& $\mathbf{0.000e+00}$& $\mathbf{0.00e+00}$&=\\
$f_{ 2 }$ & $1.098e+00$& $1.44e+00$& $\mathbf{7.244e-01}$& $\mathbf{9.34e-01}$&=\\
$f_{ 3 }$ & $\mathbf{3.219e-04}$& $\mathbf{2.87e-05}$& $1.061e-02$& $3.99e-02$&=\\
$f_{ 4 }$ & $\mathbf{2.778e-04}$& $\mathbf{2.75e-05}$& $2.783e-04$& $1.43e-05$&=\\
$f_{ 5 }$ & $\mathbf{3.248e-03}$& $\mathbf{2.65e-04}$& $9.434e-03$& $2.41e-02$&=\\
$f_{ 6 }$ & $-4.009e+00$& $2.74e-01$& $\mathbf{-4.104e+00}$& $\mathbf{1.20e-01}$&=\\
$f_{ 7 }$ & $\mathbf{2.085e+03}$& $\mathbf{0.00e+00}$& $\mathbf{2.085e+03}$& $\mathbf{0.00e+00}$&=\\
$f_{ 8 }$ & $1.686e-05$& $2.14e-05$& $\mathbf{1.583e-05}$& $\mathbf{7.27e-06}$&=\\
$f_{ 9 }$ & $-2.000e+00$& $3.80e-04$& $\mathbf{-2.000e+00}$& $\mathbf{4.62e-05}$&=\\
$f_{ 10 }$ & $\mathbf{0.000e+00}$& $\mathbf{0.00e+00}$& $\mathbf{0.000e+00}$& $\mathbf{0.00e+00}$&=\\
$f_{ 11 }$ & $\mathbf{3.750e-07}$& $\mathbf{9.92e-07}$& $\mathbf{3.750e-07}$& $\mathbf{1.13e-06}$&=\\
$f_{ 12 }$ & $\mathbf{0.000e+00}$& $\mathbf{0.00e+00}$& $\mathbf{0.000e+00}$& $\mathbf{0.00e+00}$&=\\
$f_{ 13 }$ & $\mathbf{0.000e+00}$& $\mathbf{0.00e+00}$& $\mathbf{0.000e+00}$& $\mathbf{0.00e+00}$&=\\
$f_{ 14 }$ & $\mathbf{0.000e+00}$& $\mathbf{0.00e+00}$& $\mathbf{0.000e+00}$& $\mathbf{0.00e+00}$&=\\
$f_{ 15 }$ & $4.150e-06$& $5.55e-06$& $\mathbf{2.250e-06}$& $\mathbf{2.37e-06}$&=\\
\noalign{\smallskip}\hline
\end{tabular}
\vspace{-2 mm}
\end{table}
\begin{table}[!ht]
\small
\centering
\caption{Experimental results (average final value $\pm$ std. dev. and Wilcoxon test, reference DOWSN-SA with 5 nodes) in 15 dimensions. Homogeneous DOWSN networks with 5 and 10 nodes.}
\label{tab:nr_nodes15}
\setlength{\tabcolsep}{.5em}
\begin{tabular}{c|r@{$\,\pm\,$}l|r@{$\,\pm\,$}l|c}
\hline\noalign{\smallskip}
\# & \multicolumn{2}{c|}{DOWSN-SA (5 nodes)} & \multicolumn{2}{c|}{DOWSN-SA (10 nodes)} & W \\
\noalign{\smallskip}\hline\noalign{\smallskip}
$f_{ 1 }$ & $9.282e-04$& $9.40e-04$& $\mathbf{7.386e-04}$& $\mathbf{7.28e-04}$&=\\
$f_{ 2 }$ & $1.180e+01$& $2.79e+00$& $\mathbf{1.069e+01}$& $\mathbf{4.33e+00}$&=\\
$f_{ 3 }$ & $\mathbf{1.284e+00}$& $\mathbf{7.79e-01}$& $1.355e+00$& $5.84e-01$&=\\
$f_{ 4 }$ & $\mathbf{2.059e-03}$& $\mathbf{2.59e-04}$& $2.158e-03$& $9.70e-05$&=\\
$f_{ 5 }$ & $\mathbf{3.580e+00}$& $\mathbf{2.10e+00}$& $5.032e+00$& $2.71e+00$&=\\
$f_{ 6 }$ & $-8.032e+00$& $6.41e-01$& $\mathbf{-8.198e+00}$& $\mathbf{4.56e-01}$&=\\
$f_{ 7 }$ & $6.256e+03$& $2.75e-01$& $\mathbf{6.256e+03}$& $\mathbf{1.38e-01}$&-\\
$f_{ 8 }$ & $3.902e-01$& $1.58e-01$& $\mathbf{3.059e-01}$& $\mathbf{1.08e-01}$&=\\
$f_{ 9 }$ & $-1.918e+00$& $6.35e-02$& $\mathbf{-1.943e+00}$& $\mathbf{4.73e-02}$&=\\
$f_{ 10 }$ & $9.664e-02$& $3.81e-02$& $\mathbf{7.116e-02}$& $\mathbf{3.07e-02}$&-\\
$f_{ 11 }$ & $5.152e-03$& $1.87e-03$& $\mathbf{4.491e-03}$& $\mathbf{1.17e-03}$&=\\
$f_{ 12 }$ & $8.994e-03$& $6.54e-03$& $\mathbf{7.261e-03}$& $\mathbf{5.53e-03}$&=\\
$f_{ 13 }$ & $3.312e-02$& $2.95e-02$& $\mathbf{2.941e-02}$& $\mathbf{1.99e-02}$&=\\
$f_{ 14 }$ & $\mathbf{0.000e+00}$& $\mathbf{0.00e+00}$& $\mathbf{0.000e+00}$& $\mathbf{0.00e+00}$&=\\
$f_{ 15 }$ & $9.451e-01$& $4.44e-01$& $\mathbf{7.341e-01}$& $\mathbf{2.49e-01}$&=\\
\noalign{\smallskip}\hline
\end{tabular}
\vspace{-2 mm}
\end{table}
\begin{table}[!ht]
\small
\centering
\caption{Experimental results (average final value $\pm$ std. dev. and Wilcoxon test, reference DOWSN-SA with 5 nodes) in 25 dimensions. Homogeneous DOWSN networks with 5 and 10 nodes.}
\label{tab:nr_nodes25}
\setlength{\tabcolsep}{.5em}
\begin{tabular}{c|r@{$\,\pm\,$}l|r@{$\,\pm\,$}l|c}
\hline\noalign{\smallskip}
\# & \multicolumn{2}{c|}{5 nodes} & \multicolumn{2}{c|}{10 nodes} & W \\
\noalign{\smallskip}\hline\noalign{\smallskip}
$f_{ 1 }$ & $1.279e-01$& $5.62e-02$& $\mathbf{1.206e-01}$& $\mathbf{3.49e-02}$&=\\
$f_{ 2 }$ & $\mathbf{1.222e+02}$& $\mathbf{2.79e+01}$& $1.350e+02$& $3.24e+01$&=\\
$f_{ 3 }$ & $\mathbf{2.838e+00}$& $\mathbf{2.63e-01}$& $2.870e+00$& $2.27e-01$&=\\
$f_{ 4 }$ & $\mathbf{5.585e-02}$& $\mathbf{1.54e-02}$& $5.901e-02$& $1.19e-02$&=\\
$f_{ 5 }$ & $4.594e+01$& $6.45e+00$& $\mathbf{4.452e+01}$& $\mathbf{3.96e+00}$&=\\
$f_{ 6 }$ & $\mathbf{-7.783e+00}$& $\mathbf{5.86e-01}$& $-7.633e+00$& $5.48e-01$&=\\
$f_{ 7 }$ & $1.044e+04$& $1.33e+00$& $\mathbf{1.044e+04}$& $\mathbf{1.26e+00}$&=\\
$f_{ 8 }$ & $4.951e+00$& $1.20e+00$& $\mathbf{4.191e+00}$& $\mathbf{5.52e-01}$&=\\
$f_{ 9 }$ & $-1.460e+00$& $2.56e-01$& $\mathbf{-1.520e+00}$& $\mathbf{2.71e-01}$&=\\
$f_{ 10 }$ & $1.085e+00$& $2.69e-01$& $\mathbf{9.343e-01}$& $\mathbf{9.61e-02}$&=\\
$f_{ 11 }$ & $3.668e-01$& $2.18e-01$& $\mathbf{3.562e-01}$& $\mathbf{1.25e-01}$&=\\
$f_{ 12 }$ & $2.156e+00$& $8.77e-01$& $\mathbf{1.528e+00}$& $\mathbf{4.73e-01}$&-\\
$f_{ 13 }$ & $1.470e+01$& $4.10e+00$& $\mathbf{1.129e+01}$& $\mathbf{5.26e+00}$&-\\
$f_{ 14 }$ & $1.190e-01$& $8.59e-02$& $\mathbf{5.836e-02}$& $\mathbf{2.65e-02}$&-\\
$f_{ 15 }$ & $1.323e+01$& $2.84e+00$& $\mathbf{1.143e+01}$& $\mathbf{2.33e+00}$&=\\
\noalign{\smallskip}\hline
\end{tabular}
\vspace{-2 mm}
\end{table}
We now analyze the effect of the network size, i.e. the number of nodes involved in the DOWSN optimization process, on the ``network'' optimization performance. Given that homogeneous networks perform better than equally-sized heterogeneous ones, and that the exchanged information among nodes is beneficial, we now focus only on the DOWSN-SA configuration. We compare the results obtained with random-topology DOWSN-SA networks composed of $5$ nodes (with imitation rate $0.9$ and communication period $0.25$ s), as reported in the previous section, with results obtained with analogous networks made up of $10$ nodes. Numerical results, obtained again with $16$ simulations per each network size and test function, are reported in Tables \ref{tab:nr_nodes05}-\ref{tab:nr_nodes25}.

It is quite clear that the influence of the network size on the overall optimization performance is quite negligible. However, while the statistical comparison between results obtained with $5$ and $10$ nodes leads systematically to a draw (except two cases in $15$ dimensions and three cases in $25$, where $10$ nodes obtain a better result), it is also evident that a higher number of nodes results in slightly best average fitness values on most of the test functions considered ($35$ out of $45$). The reason for this result might be twofold: on one hand, it is likely that a small number of nodes, five in this case, provide sufficient computational resources for tackling the benchmark under study. This is especially true for $5$-dimensional problems: for this dimensionality, the global optimum is found in six cases, even with networks with $5$ nodes. 
On the other hand, it is also plausible that with larger problems a higher number of nodes (thus more computational resources) would produce better results. 

The second reason for the similar performances lies in the communication mechanism adopted: a higher number of nodes generates a larger amount of network traffic (see Fig. \ref{fig:network_dynamics}), which is more prone to packet collisions. The higher is the number of collisions, the higher is the chance of losing some improvements exchanged among nodes. This destructive phenomenon eventually turns into a distributed optimization process whose efficacy does not necessarily grow with the number of nodes in the network. In other words it seems that DOWSN is efficient even (and especially) when a small number of nodes is employed. Interestingly to notice, a hierarchical framework based on DOWSN might be easily envisioned, where small clusters of nodes perform a cluster-local optimization process and exchange information, with a slower period, with other clusters present in the network.
\subsection{Influence of the imitation rate}\label{sec:imitation}
Another parameter for which it is interesting to study the influence on the ``network'' optimization performance is the imitation rate, i.e. the probability that an incoming individual is accepted for updating the node-local best. We focus on DOWSN-SA networks composed of $5$ nodes (with a communication period of $0.25$ s) and we compare the results obtained with different values of imitation rate, i.e. $0.1$, $0.5$, $0.9$ (the standard value used in the previous experiments) and $1.0$. Numerical results, based on $16$ simulations per each imitation rate and test function, are reported in Tables \ref{tab:imitation05}-\ref{tab:imitation25}. 
 
From the experiments it can be seen that the imitation rate affects the optimization especially on larger-dimensional problems. On $5$-dimensional problems, indeed, the results obtained with $q=0.9$ are equivalent in $9$ cases to those obtained with $q=0.1$ and $q=1.0$, and in $11$ cases to those obtained with $q=0.5$. Thus it can be concluded that in $5$ dimensions the imitation rate has a limited influence. On the other hand, in case of $15$ dimensions, the configuration with $q=0.9$ systematically obtains better results (or equivalent, in four cases) than configurations using $q=0.1$, $q=0.5$, or $q=1.0$. Similar consideration can be done for $25$ dimensions, where except eight cases of equivalence, the value $0.9$ produces the best results. In general, this value guarantees the overall best performance on the whole benchmark at different levels of dimensionality. This value seems to offer the best trade-off between a deterministic (unitary imitation rate) and an improbable (imitation rate $0.1$) 
acceptance of solution updates: the first condition likely causes an excessive exploitation and a diversity impoverishment, the second excessively promotes exploration and almost suppress, de facto, the effect of the exchange of information.
\begin{table*}[!ht]
\small
\centering
\caption{Experimental results (average final value $\pm$ std. dev. and Wilcoxon test, reference DOWSN-SA with imitation rate 0.9) in 5 dimensions. Homogeneous DOWSN networks with 5 nodes and different imitation rate values (in parentheses).}
\label{tab:imitation05}
\begin{tabular}{c|r@{$\,\pm\,$}l|r@{$\,\pm\,$}l|c|r@{$\,\pm\,$}l|c|r@{$\,\pm\,$}l|c}
\hline\noalign{\smallskip}
\# & \multicolumn{2}{c|}{DOWSN-SA (0.9)} & \multicolumn{2}{c|}{DOWSN-SA (0.1)} & W & \multicolumn{2}{c|}{DOWSN-SA (0.5)} & W & \multicolumn{2}{c|}{DOWSN-SA (1.0)} & W \\
\noalign{\smallskip}\hline\noalign{\smallskip}
$f_{ 1 }$ & $\mathbf{0.000e+00}$& $\mathbf{0.00e+00}$& $\mathbf{0.000e+00}$& $\mathbf{0.00e+00}$&=& $\mathbf{0.000e+00}$& $\mathbf{0.00e+00}$&=& $\mathbf{0.000e+00}$& $\mathbf{0.00e+00}$&=\\
$f_{ 2 }$ & $\mathbf{1.098e+00}$& $\mathbf{1.44e+00}$& $1.182e+00$& $6.79e-01$&=& $1.367e+00$& $5.15e-01$&=& $1.410e+00$& $4.93e-01$&=\\
$f_{ 3 }$ & $\mathbf{3.219e-04}$& $\mathbf{2.87e-05}$& $4.152e-02$& $1.09e-01$&+& $4.151e-02$& $1.09e-01$&=& $5.525e-03$& $1.37e-02$&+\\
$f_{ 4 }$ & $\mathbf{2.778e-04}$& $\mathbf{2.75e-05}$& $3.001e-04$& $2.67e-05$&+& $2.909e-04$& $1.27e-05$&=& $2.894e-04$& $1.44e-05$&+\\
$f_{ 5 }$ & $\mathbf{3.248e-03}$& $\mathbf{2.65e-04}$& $5.325e-02$& $8.63e-02$&+& $7.838e-02$& $1.39e-01$&+& $5.316e-02$& $8.63e-02$&+\\
$f_{ 6 }$ & $\mathbf{-4.009e+00}$& $\mathbf{2.74e-01}$& $-3.437e+00$& $1.11e-01$&+& $-3.498e+00$& $2.84e-01$&+& $-3.431e+00$& $3.23e-01$&+\\
$f_{ 7 }$ & $\mathbf{2.085e+03}$& $\mathbf{0.00e+00}$& $\mathbf{2.085e+03}$& $\mathbf{0.00e+00}$&=& $\mathbf{2.085e+03}$& $\mathbf{0.00e+00}$&=& $\mathbf{2.085e+03}$& $\mathbf{0.00e+00}$&=\\
$f_{ 8 }$ & $\mathbf{1.686e-05}$& $\mathbf{2.14e-05}$& $4.700e-05$& $2.02e-05$&+& $2.912e-05$& $8.19e-06$&+& $4.163e-05$& $2.32e-05$&+\\
$f_{ 9 }$ & $\mathbf{-2.000e+00}$& $\mathbf{3.80e-04}$& $-1.999e+00$& $2.34e-03$&+& $-2.000e+00$& $2.28e-04$&=& $-1.999e+00$& $9.69e-04$&=\\
$f_{ 10 }$ & $\mathbf{0.000e+00}$& $\mathbf{0.00e+00}$& $\mathbf{0.000e+00}$& $\mathbf{0.00e+00}$&=& $\mathbf{0.000e+00}$& $\mathbf{0.00e+00}$&=& $\mathbf{0.000e+00}$& $\mathbf{0.00e+00}$&=\\
$f_{ 11 }$ & $\mathbf{3.750e-07}$& $\mathbf{9.92e-07}$& $1.125e-06$& $1.45e-06$&=& $3.375e-06$& $2.34e-06$&+& $1.500e-06$& $1.50e-06$&=\\
$f_{ 12 }$ & $\mathbf{0.000e+00}$& $\mathbf{0.00e+00}$& $\mathbf{0.000e+00}$& $\mathbf{0.00e+00}$&=& $\mathbf{0.000e+00}$& $\mathbf{0.00e+00}$&=& $\mathbf{0.000e+00}$& $\mathbf{0.00e+00}$&=\\
$f_{ 13 }$ & $\mathbf{0.000e+00}$& $\mathbf{0.00e+00}$& $\mathbf{0.000e+00}$& $\mathbf{0.00e+00}$&=& $\mathbf{0.000e+00}$& $\mathbf{0.00e+00}$&=& $\mathbf{0.000e+00}$& $\mathbf{0.00e+00}$&=\\
$f_{ 14 }$ & $\mathbf{0.000e+00}$& $\mathbf{0.00e+00}$& $\mathbf{0.000e+00}$& $\mathbf{0.00e+00}$&=& $\mathbf{0.000e+00}$& $\mathbf{0.00e+00}$&=& $\mathbf{0.000e+00}$& $\mathbf{0.00e+00}$&=\\
$f_{ 15 }$ & $\mathbf{4.150e-06}$& $\mathbf{5.55e-06}$& $6.375e-06$& $4.85e-06$&=& $6.400e-06$& $5.35e-06$&=& $1.165e-05$& $8.30e-06$&+\\
\noalign{\smallskip}\hline
\end{tabular}
\end{table*}
\begin{table*}[!ht]
\small
\centering
\caption{Experimental results (average final value $\pm$ std. dev. and Wilcoxon test, reference DOWSN-SA with imitation rate 0.9) in 15 dimensions. Homogeneous DOWSN networks with 5 nodes and different imitation rate values (in parentheses).}
\label{tab:imitation15}
\begin{tabular}{c|r@{$\,\pm\,$}l|r@{$\,\pm\,$}l|c|r@{$\,\pm\,$}l|c|r@{$\,\pm\,$}l|c}
\hline\noalign{\smallskip}
\# & \multicolumn{2}{c|}{DOWSN-SA (0.9)} & \multicolumn{2}{c|}{DOWSN-SA (0.1)} & W & \multicolumn{2}{c|}{DOWSN-SA (0.5)} & W & \multicolumn{2}{c|}{DOWSN-SA (1.0)} & W \\
\noalign{\smallskip}\hline\noalign{\smallskip}
$f_{ 1 }$ & $\mathbf{9.282e-04}$& $\mathbf{9.40e-04}$& $3.039e-03$& $1.12e-03$&+& $3.828e-03$& $8.59e-04$&+& $4.152e-03$& $1.33e-03$&+\\
$f_{ 2 }$ & $\mathbf{1.180e+01}$& $\mathbf{2.79e+00}$& $3.461e+01$& $1.35e+01$&+& $3.777e+01$& $2.05e+01$&+& $3.247e+01$& $1.56e+01$&+\\
$f_{ 3 }$ & $\mathbf{1.284e+00}$& $\mathbf{7.79e-01}$& $2.301e+00$& $3.65e-01$&+& $1.885e+00$& $4.90e-01$&=& $2.011e+00$& $3.97e-01$&+\\
$f_{ 4 }$ & $\mathbf{2.059e-03}$& $\mathbf{2.59e-04}$& $2.446e-03$& $2.33e-04$&+& $2.474e-03$& $2.83e-04$&+& $2.434e-03$& $2.13e-04$&+\\
$f_{ 5 }$ & $\mathbf{3.580e+00}$& $\mathbf{2.10e+00}$& $1.008e+01$& $5.65e+00$&+& $8.358e+00$& $3.62e+00$&+& $6.247e+00$& $2.88e+00$&+\\
$f_{ 6 }$ & $\mathbf{-8.032e+00}$& $\mathbf{6.41e-01}$& $-7.248e+00$& $5.52e-01$&+& $-7.443e+00$& $3.21e-01$&+& $-7.375e+00$& $4.45e-01$&+\\
$f_{ 7 }$ & $\mathbf{6.256e+03}$& $\mathbf{2.75e-01}$& $6.256e+03$& $3.39e-01$&+& $6.256e+03$& $2.40e-01$&+& $6.256e+03$& $3.56e-01$&+\\
$f_{ 8 }$ & $\mathbf{3.902e-01}$& $\mathbf{1.58e-01}$& $1.036e+00$& $1.63e-01$&+& $9.051e-01$& $9.50e-02$&+& $9.857e-01$& $1.90e-01$&+\\
$f_{ 9 }$ & $\mathbf{-1.918e+00}$& $\mathbf{6.35e-02}$& $-1.644e+00$& $1.27e-01$&+& $-1.691e+00$& $1.73e-01$&+& $-1.644e+00$& $1.55e-01$&+\\
$f_{ 10 }$ & $\mathbf{9.664e-02}$& $\mathbf{3.81e-02}$& $1.762e-01$& $3.29e-02$&+& $1.708e-01$& $3.22e-02$&+& $1.663e-01$& $4.13e-02$&+\\
$f_{ 11 }$ & $\mathbf{5.152e-03}$& $\mathbf{1.87e-03}$& $7.970e-03$& $1.28e-03$&+& $8.412e-03$& $1.14e-03$&+& $8.548e-03$& $7.36e-04$&+\\
$f_{ 12 }$ & $\mathbf{8.994e-03}$& $\mathbf{6.54e-03}$& $4.119e-02$& $2.25e-02$&+& $3.312e-02$& $1.92e-02$&+& $3.220e-02$& $1.82e-02$&+\\
$f_{ 13 }$ & $\mathbf{3.312e-02}$& $\mathbf{2.95e-02}$& $2.294e-01$& $1.42e-01$&+& $1.102e-01$& $6.43e-02$&+& $1.898e-01$& $5.00e-02$&+\\
$f_{ 14 }$ & $\mathbf{0.000e+00}$& $\mathbf{0.00e+00}$& $5.700e-06$& $1.20e-05$&=& $4.575e-06$& $1.21e-05$&=& $\mathbf{0.000e+00}$& $\mathbf{0.00e+00}$&=\\
$f_{ 15 }$ & $\mathbf{9.451e-01}$& $\mathbf{4.44e-01}$& $2.136e+00$& $3.09e-01$&+& $2.531e+00$& $3.63e-01$&+& $2.296e+00$& $3.48e-01$&+\\
\noalign{\smallskip}\hline
\end{tabular}
\end{table*}
\begin{table*}[!ht]
\small
\centering
\caption{Experimental results (average final value $\pm$ std. dev. and Wilcoxon test, reference DOWSN-SA with imitation rate 0.9) in 25 dimensions. Homogeneous DOWSN networks with 5 nodes and different imitation rate values (in parentheses).}
\label{tab:imitation25}
\begin{tabular}{c|r@{$\,\pm\,$}l|r@{$\,\pm\,$}l|c|r@{$\,\pm\,$}l|c|r@{$\,\pm\,$}l|c}
\hline\noalign{\smallskip}
\# & \multicolumn{2}{c|}{DOWSN-SA (0.9)} & \multicolumn{2}{c|}{DOWSN-SA (0.1)} & W & \multicolumn{2}{c|}{DOWSN-SA (0.5)} & W & \multicolumn{2}{c|}{DOWSN-SA (1.0)} & W \\
\noalign{\smallskip}\hline\noalign{\smallskip}
$f_{ 1 }$ & $\mathbf{1.279e-01}$& $\mathbf{5.62e-02}$& $1.991e-01$& $1.10e-01$&=& $2.260e-01$& $9.00e-02$&+& $2.476e-01$& $8.51e-02$&+\\
$f_{ 2 }$ & $\mathbf{1.222e+02}$& $\mathbf{2.79e+01}$& $1.739e+02$& $2.57e+01$&+& $1.923e+02$& $2.15e+01$&+& $1.924e+02$& $3.32e+01$&+\\
$f_{ 3 }$ & $\mathbf{2.838e+00}$& $\mathbf{2.63e-01}$& $3.191e+00$& $3.32e-01$&+& $3.222e+00$& $1.38e-01$&+& $3.369e+00$& $1.87e-01$&+\\
$f_{ 4 }$ & $\mathbf{5.585e-02}$& $\mathbf{1.54e-02}$& $7.332e-02$& $1.33e-02$&+& $7.193e-02$& $1.08e-02$&+& $6.479e-02$& $7.27e-03$&=\\
$f_{ 5 }$ & $\mathbf{4.594e+01}$& $\mathbf{6.45e+00}$& $5.202e+01$& $5.95e+00$&=& $4.999e+01$& $4.93e+00$&=& $5.351e+01$& $4.42e+00$&+\\
$f_{ 6 }$ & $\mathbf{-7.783e+00}$& $\mathbf{5.86e-01}$& $-7.662e+00$& $3.74e-01$&=& $-7.271e+00$& $3.94e-01$&=& $-7.293e+00$& $7.92e-01$&=\\
$f_{ 7 }$ & $\mathbf{1.044e+04}$& $\mathbf{1.33e+00}$& $1.044e+04$& $1.69e+00$&+& $1.044e+04$& $1.41e+00$&+& $1.044e+04$& $1.31e+00$&+\\
$f_{ 8 }$ & $\mathbf{4.951e+00}$& $\mathbf{1.20e+00}$& $6.745e+00$& $8.81e-01$&+& $6.698e+00$& $7.60e-01$&+& $6.738e+00$& $8.27e-01$&+\\
$f_{ 9 }$ & $\mathbf{-1.460e+00}$& $\mathbf{2.56e-01}$& $-6.333e-01$& $1.71e-01$&+& $-5.137e-01$& $2.05e-01$&+& $-5.454e-01$& $2.31e-01$&+\\
$f_{ 10 }$ & $\mathbf{1.085e+00}$& $\mathbf{2.69e-01}$& $1.440e+00$& $1.84e-01$&+& $1.331e+00$& $1.07e-01$&+& $1.375e+00$& $3.16e-01$&+\\
$f_{ 11 }$ & $\mathbf{3.668e-01}$& $\mathbf{2.18e-01}$& $6.014e-01$& $1.92e-01$&+& $6.668e-01$& $1.10e-01$&+& $6.709e-01$& $1.63e-01$&+\\
$f_{ 12 }$ & $\mathbf{2.156e+00}$& $\mathbf{8.77e-01}$& $3.746e+00$& $1.05e+00$&+& $5.118e+00$& $7.65e-01$&+& $3.627e+00$& $2.04e+00$&=\\
$f_{ 13 }$ & $\mathbf{1.470e+01}$& $\mathbf{4.10e+00}$& $2.313e+01$& $8.68e+00$&+& $2.252e+01$& $4.71e+00$&+& $2.481e+01$& $7.15e+00$&+\\
$f_{ 14 }$ & $\mathbf{1.190e-01}$& $\mathbf{8.59e-02}$& $8.283e-01$& $5.37e-01$&+& $4.290e-01$& $4.01e-01$&+& $8.340e-01$& $8.64e-01$&+\\
$f_{ 15 }$ & $\mathbf{1.323e+01}$& $\mathbf{2.84e+00}$& $1.699e+01$& $1.41e+00$&+& $1.662e+01$& $2.57e+00$&+& $1.574e+01$& $1.44e+00$&+\\
\noalign{\smallskip}\hline
\end{tabular}
\end{table*}
\subsection{Influence of the communication period}\label{sec:period}
Finally, we focus on the influence of the communication period on the ``network'' optimization performance. We analyze DOWSN-SA networks composed of $5$ nodes (with an imitation rate of $0.9$) and we compare the results obtained with different communication periods, namely $0.125$, $0.25$ (the standard value used in the previous experiments), $0.5$ and $1.0$ seconds. Numerical results, based on $16$ simulations per each communication period and test function, are reported in Tables \ref{tab:com_period05}-\ref{tab:com_period25}.

In this case there is no clear statistical evidence on which communication period produces the best optimization results. On the contrary, this parameter seems not to have any influence on the optimization performance: regardless the communication period, the DOWSN-SA configuration is always able to obtain the same (i.e. statistically equivalent) results, see the Wilcoxon tests in Tables \ref{tab:com_period05}-\ref{tab:com_period25}. This is an interesting finding as it implies that, given a computational budget sufficiently larger than the communication period (in our case $60$ seconds), even a low packet frequency is able to produce an overall good optimization performance. In other words, only a few packets exchanged during the optimization process are enough to obtain a fitness improvement on all the nodes in the network (and thus a better ``network'' average fitness). Of course a higher packet frequency is likely to produce a faster convergence on all the nodes, however transmitting a higher number of 
packets requires a higher energy consumption (due to more network system calls, which are the most power-hungry operations on WSN nodes, see next section). Based on the numerical results reported in Tables \ref{tab:com_period05}-\ref{tab:com_period25}, we believe that a communication period of $0.25$ seconds represent a fair compromise between energy consumption and information exchange. It should be noted that similar results, not reported here for the sake of brevity, can be obtained also for the other network configurations (that is, different network sizes and imitation rates). 
\begin{table*}[!ht]
\small
\centering
\caption{Experimental results (average final value $\pm$ std. dev. and Wilcoxon test, reference DOWSN-SA with communication period 0.25 s) in 5 dimensions. Homogeneous DOWSN networks with 5 nodes and different communication periods (in parentheses).}
\label{tab:com_period05}
\begin{tabular}{c|r@{$\,\pm\,$}l|r@{$\,\pm\,$}l|c|r@{$\,\pm\,$}l|c|r@{$\,\pm\,$}l|c}
\hline\noalign{\smallskip}
\# & \multicolumn{2}{c|}{DOWSN-SA (0.25 s)} & \multicolumn{2}{c|}{DOWSN-SA (1 s)} & W & \multicolumn{2}{c|}{DOWSN-SA (0.5 s)} & W & \multicolumn{2}{c|}{DOWSN-SA (0.125 s)} & W \\
\noalign{\smallskip}\hline\noalign{\smallskip}
$f_{ 1 }$ & $\mathbf{0.000e+00}$& $\mathbf{0.00e+00}$& $\mathbf{0.000e+00}$& $\mathbf{0.00e+00}$&=& $\mathbf{0.000e+00}$& $\mathbf{0.00e+00}$&=& $\mathbf{0.000e+00}$& $\mathbf{0.00e+00}$&=\\
$f_{ 2 }$ & $1.098e+00$& $1.44e+00$& $\mathbf{4.692e-01}$& $\mathbf{9.76e-02}$&=& $5.723e-01$& $6.82e-01$&=& $1.160e+00$& $1.46e+00$&=\\
$f_{ 3 }$ & $3.219e-04$& $2.87e-05$& $3.064e-03$& $7.27e-03$&=& $3.279e-04$& $3.43e-05$&=& $\mathbf{3.213e-04}$& $\mathbf{2.78e-05}$&=\\
$f_{ 4 }$ & $2.778e-04$& $2.75e-05$& $2.704e-04$& $1.67e-05$&=& $2.810e-04$& $1.88e-05$&=& $\mathbf{2.692e-04}$& $\mathbf{1.55e-05}$&=\\
$f_{ 5 }$ & $3.248e-03$& $2.65e-04$& $\mathbf{3.230e-03}$& $\mathbf{1.81e-04}$&=& $2.807e-02$& $6.59e-02$&=& $2.805e-02$& $6.59e-02$&=\\
$f_{ 6 }$ & $-4.009e+00$& $2.74e-01$& $-3.853e+00$& $3.38e-01$&=& $-3.988e+00$& $1.85e-01$&=& $\mathbf{-4.021e+00}$& $\mathbf{1.74e-01}$&=\\
$f_{ 7 }$ & $\mathbf{2.085e+03}$& $\mathbf{0.00e+00}$& $\mathbf{2.085e+03}$& $\mathbf{0.00e+00}$&=& $\mathbf{2.085e+03}$& $\mathbf{0.00e+00}$&=& $\mathbf{2.085e+03}$& $\mathbf{0.00e+00}$&=\\
$f_{ 8 }$ & $1.686e-05$& $2.14e-05$& $\mathbf{6.400e-06}$& $\mathbf{8.99e-06}$&=& $2.080e-05$& $2.34e-05$&=& $9.750e-06$& $1.05e-05$&=\\
$f_{ 9 }$ & $-2.000e+00$& $3.80e-04$& $\mathbf{-2.000e+00}$& $\mathbf{1.95e-05}$&=& $-2.000e+00$& $3.45e-05$&=& $-2.000e+00$& $5.73e-05$&=\\
$f_{ 10 }$ & $\mathbf{0.000e+00}$& $\mathbf{0.00e+00}$& $\mathbf{0.000e+00}$& $\mathbf{0.00e+00}$&=& $\mathbf{0.000e+00}$& $\mathbf{0.00e+00}$&=& $\mathbf{0.000e+00}$& $\mathbf{0.00e+00}$&=\\
$f_{ 11 }$ & $\mathbf{3.750e-07}$& $\mathbf{9.92e-07}$& $2.250e-06$& $3.27e-06$&=& $\mathbf{3.750e-07}$& $\mathbf{9.92e-07}$&=& $\mathbf{3.750e-07}$& $\mathbf{9.92e-07}$&=\\
$f_{ 12 }$ & $\mathbf{0.000e+00}$& $\mathbf{0.00e+00}$& $\mathbf{0.000e+00}$& $\mathbf{0.00e+00}$&=& $\mathbf{0.000e+00}$& $\mathbf{0.00e+00}$&=& $\mathbf{0.000e+00}$& $\mathbf{0.00e+00}$&=\\
$f_{ 13 }$ & $\mathbf{0.000e+00}$& $\mathbf{0.00e+00}$& $\mathbf{0.000e+00}$& $\mathbf{0.00e+00}$&=& $\mathbf{0.000e+00}$& $\mathbf{0.00e+00}$&=& $\mathbf{0.000e+00}$& $\mathbf{0.00e+00}$&=\\
$f_{ 14 }$ & $\mathbf{0.000e+00}$& $\mathbf{0.00e+00}$& $\mathbf{0.000e+00}$& $\mathbf{0.00e+00}$&=& $\mathbf{0.000e+00}$& $\mathbf{0.00e+00}$&=& $\mathbf{0.000e+00}$& $\mathbf{0.00e+00}$&=\\
$f_{ 15 }$ & $4.150e-06$& $5.55e-06$& $2.625e-06$& $3.50e-06$&=& $3.750e-06$& $5.14e-06$&=& $\mathbf{1.500e-06}$& $\mathbf{1.50e-06}$&=\\
\noalign{\smallskip}\hline
\end{tabular}
\end{table*}
\begin{table*}[!ht]
\small
\centering
\caption{Experimental results (average final value $\pm$ std. dev. and Wilcoxon test, reference DOWSN-SA with communication period 0.25 s) in 15 dimensions. Homogeneous DOWSN networks with 5 nodes and different communication periods (in parentheses).}
\label{tab:com_period15}
\begin{tabular}{c|r@{$\,\pm\,$}l|r@{$\,\pm\,$}l|c|r@{$\,\pm\,$}l|c|r@{$\,\pm\,$}l|c}
\hline\noalign{\smallskip}
\# & \multicolumn{2}{c|}{DOWSN-SA (0.25 s)} & \multicolumn{2}{c|}{DOWSN-SA (1 s)} & W & \multicolumn{2}{c|}{DOWSN-SA (0.5 s)} & W & \multicolumn{2}{c|}{DOWSN-SA (0.125 s)} & W \\
\noalign{\smallskip}\hline\noalign{\smallskip}
$f_{ 1 }$ & $\mathbf{9.282e-04}$& $\mathbf{9.40e-04}$& $9.761e-04$& $1.03e-03$&=& $1.491e-03$& $9.31e-04$&=& $1.022e-03$& $1.41e-03$&=\\
$f_{ 2 }$ & $1.180e+01$& $2.79e+00$& $\mathbf{1.029e+01}$& $\mathbf{3.99e+00}$&=& $3.597e+01$& $2.67e+01$&=& $2.640e+01$& $2.05e+01$&=\\
$f_{ 3 }$ & $\mathbf{1.284e+00}$& $\mathbf{7.79e-01}$& $1.458e+00$& $6.15e-01$&=& $1.493e+00$& $6.99e-01$&=& $1.346e+00$& $5.23e-01$&=\\
$f_{ 4 }$ & $2.059e-03$& $2.59e-04$& $\mathbf{1.997e-03}$& $\mathbf{3.19e-04}$&=& $2.229e-03$& $1.69e-04$&=& $2.028e-03$& $3.53e-04$&=\\
$f_{ 5 }$ & $3.580e+00$& $2.10e+00$& $4.235e+00$& $2.61e+00$&=& $\mathbf{3.439e+00}$& $\mathbf{2.46e+00}$&=& $4.421e+00$& $3.49e+00$&=\\
$f_{ 6 }$ & $-8.032e+00$& $6.41e-01$& $-8.292e+00$& $5.56e-01$&=& $-7.613e+00$& $5.47e-01$&=& $\mathbf{-8.449e+00}$& $\mathbf{6.25e-01}$&=\\
$f_{ 7 }$ & $6.256e+03$& $2.75e-01$& $\mathbf{6.256e+03}$& $\mathbf{1.89e-01}$&=& $6.256e+03$& $3.13e-01$&=& $6.256e+03$& $3.25e-01$&=\\
$f_{ 8 }$ & $3.902e-01$& $1.58e-01$& $3.902e-01$& $1.54e-01$&=& $\mathbf{3.497e-01}$& $\mathbf{1.97e-01}$&=& $4.962e-01$& $2.15e-01$&=\\
$f_{ 9 }$ & $\mathbf{-1.918e+00}$& $\mathbf{6.35e-02}$& $-1.861e+00$& $1.23e-01$&=& $-1.836e+00$& $1.62e-01$&=& $-1.863e+00$& $1.71e-01$&=\\
$f_{ 10 }$ & $9.664e-02$& $3.81e-02$& $8.241e-02$& $2.40e-02$&=& $\mathbf{7.296e-02}$& $\mathbf{2.49e-02}$&=& $9.209e-02$& $3.44e-02$&=\\
$f_{ 11 }$ & $\mathbf{5.152e-03}$& $\mathbf{1.87e-03}$& $6.335e-03$& $1.75e-03$&=& $5.427e-03$& $2.81e-03$&=& $5.451e-03$& $2.23e-03$&=\\
$f_{ 12 }$ & $8.994e-03$& $6.54e-03$& $\mathbf{7.194e-03}$& $\mathbf{5.24e-03}$&=& $1.117e-02$& $7.51e-03$&=& $8.760e-03$& $7.46e-03$&=\\
$f_{ 13 }$ & $3.312e-02$& $2.95e-02$& $\mathbf{2.421e-02}$& $\mathbf{2.13e-02}$&=& $3.501e-02$& $3.30e-02$&=& $2.607e-02$& $1.78e-02$&=\\
$f_{ 14 }$ & $\mathbf{0.000e+00}$& $\mathbf{0.00e+00}$& $\mathbf{0.000e+00}$& $\mathbf{0.00e+00}$&=& $\mathbf{0.000e+00}$& $\mathbf{0.00e+00}$&=& $\mathbf{0.000e+00}$& $\mathbf{0.00e+00}$&=\\
$f_{ 15 }$ & $9.451e-01$& $4.44e-01$& $8.316e-01$& $2.74e-01$&=& $1.000e+00$& $3.60e-01$&=& $\mathbf{7.443e-01}$& $\mathbf{1.04e-01}$&=\\
\noalign{\smallskip}\hline
\end{tabular}
\end{table*}
\begin{table*}[!ht]
\small
\centering
\caption{Experimental results (average final value $\pm$ std. dev. and Wilcoxon test, reference DOWSN-SA with communication period 0.25 s) in 25 dimensions. Homogeneous DOWSN networks with 5 nodes and different communication periods (in parentheses).}
\label{tab:com_period25}
\begin{tabular}{c|r@{$\,\pm\,$}l|r@{$\,\pm\,$}l|c|r@{$\,\pm\,$}l|c|r@{$\,\pm\,$}l|c}
\hline\noalign{\smallskip}
\# & \multicolumn{2}{c|}{DOWSN-SA (0.25 s)} & \multicolumn{2}{c|}{DOWSN-SA (1 s)} & W & \multicolumn{2}{c|}{DOWSN-SA (0.5 s)} & W & \multicolumn{2}{c|}{DOWSN-SA (0.125 s)} & W \\
\noalign{\smallskip}\hline\noalign{\smallskip}
$f_{ 1 }$ & $1.279e-01$& $5.62e-02$& $1.192e-01$& $4.75e-02$&=& $1.638e-01$& $7.43e-02$&=& $\mathbf{1.144e-01}$& $\mathbf{3.45e-02}$&=\\
$f_{ 2 }$ & $\mathbf{1.222e+02}$& $\mathbf{2.79e+01}$& $1.276e+02$& $3.36e+01$&=& $1.486e+02$& $4.67e+01$&=& $1.235e+02$& $3.56e+01$&=\\
$f_{ 3 }$ & $\mathbf{2.838e+00}$& $\mathbf{2.63e-01}$& $2.927e+00$& $1.85e-01$&=& $2.952e+00$& $3.91e-01$&=& $2.917e+00$& $2.95e-01$&=\\
$f_{ 4 }$ & $\mathbf{5.585e-02}$& $\mathbf{1.54e-02}$& $5.950e-02$& $1.55e-02$&=& $5.646e-02$& $2.43e-02$&=& $6.036e-02$& $1.83e-02$&=\\
$f_{ 5 }$ & $4.594e+01$& $6.45e+00$& $\mathbf{4.441e+01}$& $\mathbf{5.19e+00}$&=& $4.868e+01$& $7.86e+00$&=& $4.511e+01$& $6.02e+00$&=\\
$f_{ 6 }$ & $\mathbf{-7.783e+00}$& $\mathbf{5.86e-01}$& $-7.374e+00$& $5.44e-01$&=& $-7.496e+00$& $4.85e-01$&=& $-7.765e+00$& $5.44e-01$&=\\
$f_{ 7 }$ & $1.044e+04$& $1.33e+00$& $1.044e+04$& $1.58e+00$&=& $\mathbf{1.044e+04}$& $\mathbf{1.30e+00}$&=& $1.044e+04$& $1.36e+00$&=\\
$f_{ 8 }$ & $4.951e+00$& $1.20e+00$& $\mathbf{4.197e+00}$& $\mathbf{8.14e-01}$&=& $4.577e+00$& $1.03e+00$&=& $4.438e+00$& $1.33e+00$&=\\
$f_{ 9 }$ & $-1.460e+00$& $2.56e-01$& $-1.516e+00$& $2.20e-01$&=& $\mathbf{-1.545e+00}$& $\mathbf{1.60e-01}$&=& $-1.477e+00$& $3.57e-01$&=\\
$f_{ 10 }$ & $1.085e+00$& $2.69e-01$& $\mathbf{9.101e-01}$& $\mathbf{1.92e-01}$&=& $9.360e-01$& $1.08e-01$&=& $9.545e-01$& $1.42e-01$&=\\
$f_{ 11 }$ & $3.668e-01$& $2.18e-01$& $4.053e-01$& $1.47e-01$&=& $4.502e-01$& $1.40e-01$&=& $\mathbf{2.769e-01}$& $\mathbf{1.25e-01}$&=\\
$f_{ 12 }$ & $2.156e+00$& $8.77e-01$& $\mathbf{1.984e+00}$& $\mathbf{8.08e-01}$&=& $2.443e+00$& $1.08e+00$&=& $2.040e+00$& $1.16e+00$&=\\
$f_{ 13 }$ & $1.470e+01$& $4.10e+00$& $1.571e+01$& $7.38e+00$&=& $\mathbf{1.193e+01}$& $\mathbf{3.55e+00}$&=& $1.572e+01$& $4.20e+00$&=\\
$f_{ 14 }$ & $\mathbf{1.190e-01}$& $\mathbf{8.59e-02}$& $2.240e-01$& $2.86e-01$&=& $2.225e-01$& $2.19e-01$&=& $3.338e-01$& $3.18e-01$&=\\
$f_{ 15 }$ & $1.323e+01$& $2.84e+00$& $\mathbf{1.104e+01}$& $\mathbf{2.58e+00}$&=& $1.180e+01$& $2.77e+00$&=& $1.223e+01$& $2.96e+00$&=\\
\noalign{\smallskip}\hline
\end{tabular}
\end{table*}
\section{Hardware resource usage}\label{sec:hardware}
As we have seen at the beginning of the paper, wireless sensors are severely limited especially in terms of memory and energy. We here conclude our analysis focusing on how DOWSN deals with these limitations and how hardware resources are used during the optimization. Again we focus the analysis on the TelosB mote family.

\subsection{Energy consumption}\label{sec:power}
TelosB motes can operate in various states \cite{bib:telos}, depending on the operating conditions of the MSP430 Micro-Controller Unit (MCU), which can work in full or low power mode, and the CC2420 transceiver, which in turn can be in a receive (listen, RX) or send (transmit, TX) mode. Apart from the MCU standby modality, four working states can be identified, namely \emph{cpu} (MCU on, radio off), \emph{rx} (MCU on, radio RX), \emph{tx} (MCU on, radio TX), \emph{lpm} (Low Power Mode: MCU idle, radio off). The latter state uses approximately $3\%$ of full power and is activated only when there isn't any process running or event pending. For each of the four states, the current drawn by the mote can be experimentally measured, or found on TelosB data sheets. According to \cite{bib:telos}, operational currents are approximately $1.8$ mA for \emph{cpu} state, $19.5$ mA and $21.8$ mA respectively for \emph{tx} and \emph{rx} modes, and $54.5$ $\mu$A for the low power mode.

Contiki provides a specific API (``energest'') for measuring the amount of time a mote spends in each state, in real-time ticks. Thus the power consumption on each mote can be profiled simply multiplying each time frame by the corresponding operational current. In our experience, this API is used to log periodically the above time frames. These data are then post-processed in Python for further analysis.

\begin{figure*}[!ht]
\centering
\includegraphics[width=0.75\textwidth]{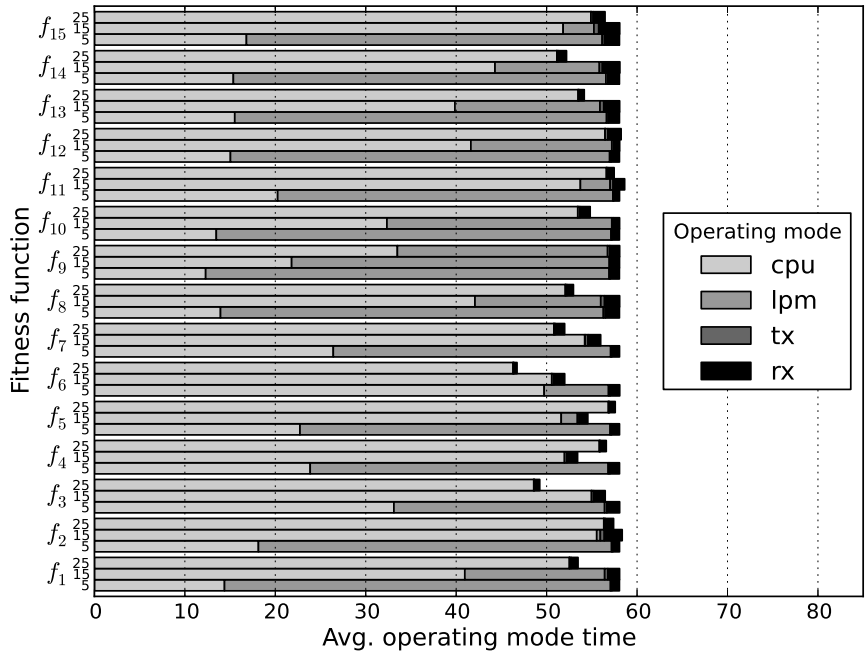}
\includegraphics[width=0.75\textwidth]{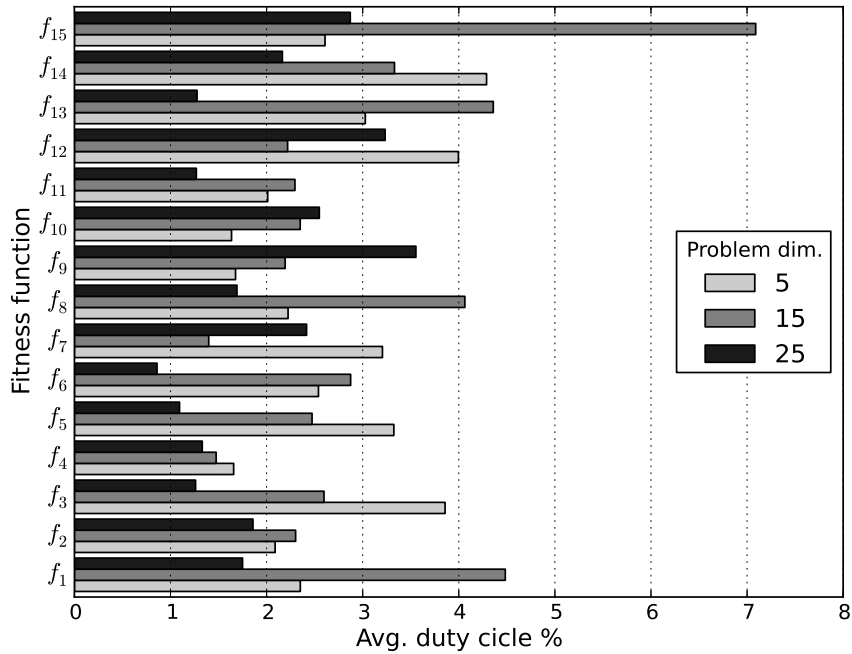}
\caption{Energetic behaviour, in dependence on problem and dimension, of DOWSN-SA configurations with $5$ nodes, imitation rate $0.9$ and communication period $0.25$ s}
\label{fig:power}
\end{figure*}

Fig. \ref{fig:power} (top) shows, for each benchmark function and problem size ($5$, $15$, $25$), the mote operating mode time averaged over $16$ simulations of DOWSN-SA configurations with $5$ nodes, imitation rate $0.9$ and communication period $0.25$ s. We must remark that the cumulative time does not match the computational budget ($60$ s) due to small latencies that are not accounted for during the simulation. It can be easily seen that the amount of time which motes spend, on average, in \emph{cpu} state clearly depends on the problem dimension. This is true for all the test functions and it is an obvious consequence of the increasing complexity of fitness evaluations. On the other hand, the duration of the \emph{lpm} state follows a dual trend, being longer for 5-dimensional problems and increasing for higher dimensionalities. Finally, the amount of time the radio transceiver is on (\emph{tx} and \emph{rx} modes) is for every test function and problem dimension 
(with the exception of $f_{15}$ in $15$ dimensions) in the order of $1-4.5\%$ of the cumulative time. This percentage is the mote duty cycle (being $T_{s}$ the time spent in state $s$):
\begin{equation}
duty~cycle = \frac{T_{tx} + T_{rx}}{T_{lpm} + T_{cpu} + T_{lpm} + T_{cpu}}
\end{equation}
which is shown in Fig. \ref{fig:power} (bottom) in dependence on problem and dimension. An aggregate information on the average operating mode times and duty cycles over the whole benchmark, at the three dimensionalities considered, is reported in Tab. \ref{tab:power}.a, where again it can be seen that the average duty cycle barely exceeds the $3\%$ of the total time. This result is, from a systems standpoint, very important as it indicates that the communication features of the network are used parsimoniously with respect to the global computation time. Since the \emph{tx}/\emph{rx} states are the most power-hungry, it is indeed crucial for the network lifetime to reduce their usage as much as possible. As shown in Tab. \ref{tab:power}.b, where an operational voltage of $3$ V is assumed, the average power consumption for a 5-dimensional problem is in the order of $3.4$ mW, while it is almost twice this amount for problems in $15$ and $25$ dimensions.
\begin{table}
\centering
\caption{(a) Average node operating mode time (s) and duty cycle (\%) and (b) average node power (mW) and energy (mJ) consumption in dependence on problem dimension of DOWSN-SA with $5$ nodes, imitation rate $0.9$ and communication period $0.25$ s}
\label{tab:power}
\subtable[]
{
\begin{tabular}{c|c|c|c|c||c}
\hline\noalign{\smallskip}
$n$ & $T_{cpu}$ & $T_{lpm}$ & $T_{tx}$ & $T_{rx}$ & $duty~cycle$\\
\noalign{\smallskip}\hline\noalign{\smallskip}
$5$& $20.7$& $36.1$& $0.169$& $1.06$& $2.7\%$\\
$15$& $45.8$& $9.43$& $0.308$& $1.37$& $3.03\%$\\
$25$& $51.9$& $1.55$& $0.165$& $0.851$& $1.95\%$\\
\noalign{\smallskip}\hline
\end{tabular}
}
\subtable[]
{
\begin{tabular}{c|c|c}
\hline\noalign{\smallskip}
n & Power & Energy \\
\noalign{\smallskip}\hline\noalign{\smallskip}
$5$& $3.4$& $197$\\
$15$& $6.26$& $356$\\
$25$& $6.35$& $346$\\
\noalign{\smallskip}\hline
\end{tabular}
}
\end{table}
\subsection{Memory footprint}\label{sec:memory}
In order to profile the memory consumption of DOWSN, we use the \texttt{msp430-size} utility available with the GCC toolchain for the MSP430 processor. Given a binary program file (the one that is flashed on the mote's program memory), this tool provides detailed information about the memory size of the program instructions section (\textit{text}), the initialized static data segment (\textit{data}), and the uninitialized static data segment (\textit{bss}). It must be noticed that the total process size (\textit{dec} = \textit{text} + \textit{data} + \textit{bss}), computed at compile time, does not take into account dynamical allocations (which take place instead in the \textit{heap} segment). To limit the overall memory footprint, no dynamical allocations are used in DOWSN. Additionally, DOWSN is compiled with GCC optimization level \texttt{-Os}, which performs specific compiling optimizations designed to reduce code size.
 
As reported in our previous study \cite{bib:DOWSN-UKCI}, the \textit{data} memory section of a DOWSN mote program occupies about $150$ bytes. The size of the \textit{bss} varies approximately from and $5.7$ to $6.2$ kB, depending on the problem dimensionality (from $5$ to $25$). It should be noted that this memory consumption is only $300-800$ bytes larger than the footprint of a ``dummy" Contiki program implementing an infinite empty loop. In other words, DOWSN has a very limited overhead in terms of uninitialized static data, mostly arrays encoding problem solutions (``slots'') and other preallocated data.

As for the \textit{text} section, with reference to the DOWSN node-level architecture described in section \ref{sec:dowsn}, the basic Contiki installation occupies approximately $19.8$ kB, while the network thread requires about $5.3$ kB. The memory footprint of the fitness functions considered in this study varies roughly from $2.1$ kB (Schwefel problem 2.22) to $4.7$ kB (Ackley function): this value depends mainly on the symbols which are linked from the \texttt{libfixmath} library and the additional fixed-point functions. Finally, the overhead due to the optimization algorithm obviously depends on the DOWSN configuration employed. In case of DOWSN-MA, where the whole A-DB needs to be stored in memory, this additional memory requirement is in the order of $5$ kB ($500$ bytes for RS, $800$ bytes for ISPO, $1.4$ kB for nuSA and $2.3$ kB for 3SOME). Instead a DOWSN-SA  configuration employing 3SOME only requires $2.3$ additional kB. Thus the advantage of using DOWSN-SA configurations is evident not only in 
terms of performance, but also in terms of memory.

In summary, the total memory footprint of DOWSN is in the order of $30-37$ kB, depending on the fitness function, its dimensionality, and the algorithm(s) employed. Considering that the TelosB platform has a $48$ kB program memory, $11-16$ kB are thus available for user applications. A further reduction of $8$ kB can be obtained suppressing the debug messages generated during DOWSN execution.
\section{Conclusion}\label{sec:conclusion}
In this paper we presented an extensive experimental campaign on DOWSN, a distributed optimization framework for Wireless Sensor Networks originally proposed in our previous work \cite{bib:DOWSN-UKCI}. DOWSN consists of an island model infrastructure in which each node executes an optimization algorithm and exchanges, periodically, promising solutions with its neighbors. Multiple configurations have been investigated, including heterogeneous networks, where all nodes use different algorithms, and homogeneous ones, where all nodes use the same algorithm. A selection of \emph{memory-saving} algorithms has been implemented on Contiki, an Operating System specifically designed for the embedded devices used in WSNs. Numerical simulations have been performed using the Contiki network simulator COOJA to test the optimization performance of DOWSN. The performance of DOWSN has been assessed, for three problem sizes ($5$, $15$ and $25$), on a benchmark consisting of $15$ test functions. The influence of the main 
properties of DOWSN has also been analyzed, namely the network size, the inter-node communication period, and the \emph{imitation rate}, a factor influencing the probability a node accepts incoming promising solutions. 
The main finding of this experimental campaign was that, compared to all the other configurations under investigation, a homogeneous DOWSN configuration (DOWSN-SA) composed of $5$ nodes employing the single-solution algorithm proposed in \cite{bib:Iacca20123SOME}, with an imitation rate of $0.9$ and a communication period of $0.25$ s, shows on average the best optimization results. To complement our discussion, we then performed an accurate profiling of the energy and memory consumption of DOWSN, that showed the efficient usage of the scarce resources available on the nodes. 

Future studies will further extend the proposed architecture in various ways and from different perspectives. From an algorithmic side, new optimization methods can be investigated in order to outperform the current DOWSN structure. From a network viewpoint, it would be interesting to study the influence of the network topology on the optimization performance, for example implementing a hierarchical network topology where multiple clusters of nodes perform cluster-local optimization processes and exchange information among them; additionally, alternative communication schemes could be employed instead of local broadcast. Another feature that might be investigated is a self-adapting scheme for the communication period and the imitation rate, in order to improve the network energy consumption. As for the implementation, the platform could be ported to other Operating Systems, e.g. TinyOS, and tested on different mote families where the energy consumption might be different. Finally, from an engineering 
standpoint, possible applications of DOWSN will be studied, for example in the context of distributed modelling, self-adaptation of node internal behaviour, node localization, optimal scheduling of measurements, and protocol optimization.
\begin{appendix}
\section{Appendix: Test Problems}\label{sec:test_desc}
The test problems listed in Table \ref{tab:benchmark} have been considered in this study. For each problem, the decision space was set to $\mathbf{D}=\left[-2.0,2.0\right]^{n}$, where $n$ is the problem dimensionality. As shown in Table \ref{tab:benchmark}, the testbed is composed of $15$ fitness functions with different properties in terms of modality and separability.
\begin{table*}[!ht]
\centering
\caption{Test Problems}
\renewcommand{\arraystretch}{1.5}
\label{tab:benchmark}
\begin{tabular}{c|c|c|c|c}
\hline\noalign{\smallskip}
\# & Function [Ref.] & Function Definition & Unimodal & Separable \\
\noalign{\smallskip}\hline\noalign{\smallskip}
$f_1$ & Sphere \cite{bib:Suganthan2005} & $\sum_{i=1}^{n}{x_i^2}$ & y & y \\
$f_2$ & Rosenbrock \cite{bib:Qin2009} & $\sum_{i=1}^{n-1}\left[100 \left(x_{i+1}-x_i^2\right)^2+\left(1-x_i\right)^2\right]$ & n & n \\
$f_3$ & Ackley \cite{bib:Qin2009} & $-20 e^{-0.2 \sqrt{1/n\sum_{i=1}^{n}x_i^2}} - e^{\left(1/n\right)\sum_{i=1}^{n}{\cos(2\pi x_i)}} + 20 + e$ 
      & n & n \\
$f_4$ & Griewank \cite{bib:Qin2009} & $\sum_{i=1}^{n}\frac{x_i^2}{4000}-\prod_{i=1}^{n}\cos\left(\frac{x_i}{\sqrt{i}}\right)+1$ & n & n \\
$f_5$ & Rastrigin \cite{bib:Suganthan2005} & $10n+\sum_{i=1}^{n}\left[x_i^2-10\cos\left(2\pi x_i\right)\right]$ & n & y \\
$f_6$ & Michalewicz \cite{bib:Michalewicz} & $-\sum_{i=1}^{n}\sin\left(x_i\right){\left[\sin{\left(\frac{ix_i^2}{\pi}\right)}\right]}^{20}$ 
      & n & y \\
$f_7$ & Schwefel \cite{bib:Qin2009} & $418.9829n + \sum_{i=1}^{n}\left[-x_i \sin\left(\sqrt{\left|x_i\right|}\right)\right]$ & n & y \\
$f_8$ & Schwefel 1.2 \cite{bib:Suganthan2005} & $\sum_{i=1}^{n}\left(\sum_{j=1}^{i}x_j\right)^2$ & y & n \\
$f_9$ & Schwefel 2.21 \cite{bib:Vestersrtom2004} & $\max_{i}x_i$ & y & n \\ 
$f_{10}$ & Schwefel 2.22 \cite{bib:Vestersrtom2004} & $\sum_{i=1}^{n}\left|x_i\right| + \prod_{i=1}^{n}\left|x_i\right|$ & y & y \\
$f_{11}$ & Alpine \cite{web:clerc} & $\sum_{i=1}^{n}\left|x_i\sin\left(x_i\right) + 0.1x_i \right|$ & n & y \\
$f_{12}$ & Axis Parallel \cite{bib:molga-smutnicki} & $\sum_{i=1}^{n}\left(ix_i^2\right)$ & y & n \\
$f_{13}$ & Moved Axis Parallel \cite{bib:Yap2011} & $\sum_{i=1}^{n}\left(5i x_i^2\right)$ & y & n \\
$f_{14}$ & Power Sum \cite{bib:molga-smutnicki} & $\sum_{i=1}^{n}\left(\left|x_i\right|^{i+1}\right)$ & y & n \\
$f_{15}$ & Zakharov \cite{bib:sahps} & $\sum_{i=1}^{n}x_i^2 + \left(\sum_{i=1}^{n}0.5ix_i\right)^2 + \left(\sum_{i=1}^{n}0.5ix_i\right)^4$ 
	 & y & n \\
\noalign{\smallskip}\hline
\end{tabular}
\end{table*}
\end{appendix}
\begin{acknowledgements}
INCAS\textsuperscript{3} is co-funded by the Province of Drenthe, the Municipality of Assen, the European Fund for Regional Development and the Ministry of Economic Affairs, Peaks in the Delta.
\end{acknowledgements}
\bibliographystyle{spmpsci}	

%
\end{document}